\newif\ifreview

\PassOptionsToPackage{usenames,dvipsnames,table}{xcolor}
\PassOptionsToPackage{colorlinks=purple,
  linkcolor=purple,
  citecolor=purple,
  urlcolor=purple,
  pdfauthor={},
  pdftitle={},
  pdfsubject={},
  pdfkeywords={},
  bookmarks=false,}{hyperref}
\documentclass[sigplan,nonacm, 10pt, screen]{acmart}

\usepackage{xcolor}
\usepackage{hyperref}

\acmYear{2024}\copyrightyear{2024}
\acmConference[Middleware '24]{}
\acmBooktitle{25th ACM/IFIP International Middleware Conference}
\acmPrice{XXX}
\acmDOI{XXX}
\acmISBN{XXX}

\setcopyright{acmlicensed}

\newif\ifnotes
\usepackage{amsmath,amsfonts}
\usepackage{graphicx}
\usepackage{textcomp}

\hypersetup{colorlinks=true}

\usepackage{caption}
\usepackage{subcaption}

\usepackage[linesnumbered,ruled,vlined]{algorithm2e}

\SetCommentSty{mycommfont}

\SetKwInOut{Parameter}{parameters}
\SetKwInOut{Initialize}{initialize}

\makeatletter
\algocf@newcommand{KwParameterXX}[1]{%comment
  \sbox\algocf@inputbox{\hbox{\KwSty{Parameter}\algocf@typo: }}%comment
  \ifthenelse{\boolean{algocf@inoutnumbered}}{\relax}{\everypar={\relax}}%comment
  {\let\\\algocf@newinput\hspace{\wd\algocf@inputbox}\hangindent=\wd\algocf@inputbox\hangafter=\wd\algocf@inputbox#1\par}%comment
  \algocf@linesnumbered%comment
}
\makeatother

\makeatletter
\algocf@newcommand{KwInitializeXX}[1]{%comment
  \sbox\algocf@inputbox{\hbox{\KwSty{Initialize}\algocf@typo: }}%comment
  \ifthenelse{\boolean{algocf@inoutnumbered}}{\relax}{\everypar={\relax}}%comment
  {\let\\\algocf@newinput\hspace{\wd\algocf@inputbox}\hangindent=\wd\algocf@inputbox\hangafter=\wd\algocf@inputbox#1\par}%comment
  \algocf@linesnumbered%comment
}
\makeatother

\usepackage{pifont}
\usepackage{amsmath}
\usepackage{mathtools}
\usepackage{xspace}
\usepackage{mathtools}

\DeclarePairedDelimiter{\floor}{\lfloor}{\rfloor}

\usepackage{flushend}

\usepackage{booktabs}
\usepackage{siunitx}
\usepackage[normalem]{ulem}
\usepackage{setspace}
\usepackage{multirow} %comment
\usepackage[commandnameprefix=ifneeded,truncate=hyphenate]{changes}

\definecolor{vrpink}{RGB}{255,0,127}
\definecolor{vrblue}{RGB}{30,144,255}
\definecolor{vrolive}{RGB}{85,107,47}
\definecolor{vrroyalblue}{RGB}{65,105,225}
\definecolor{brgreen}{RGB}{100,200,70}
\definecolor{ivsalmon}{RGB}{255,160,122}
\definecolor{vrlpink}{RGB}{255,192,203}
\definecolor{mvcol}{RGB}{5,150,25}
\newcommand{\cmark}{{\color{OliveGreen} \ding{51}}}
\newcommand{\xmark}{\color{BrickRed} \ding{55}}

\def\BibTeX{{\rm B\kern-.05em{\sc i\kern-.025em b}\kern-.08em
    T\kern-.1667em\lower.7ex\hbox{E}\kern-.125emX}}

\DeclarePairedDelimiter{\norm}{\lVert}{\rVert}
\newcommand{\pnameTitle}{\texttt{Catalyst}}
\newcommand{\pname}{\pnameTitle}

\newcommand{\fasync}{\texttt{FedAsync}}
\newcommand{\basgd}{\texttt{BASGD}}
\newcommand{\favg}{\texttt{FedAvg}}

\newcommand{\cifar}{\texttt{Cifar-10}}
\newcommand{\mnist}{\texttt{MNIST}}
\newcommand{\wikitext}{\texttt{WikiText-2}}

\begin{document}

\title{Asynchronous Byzantine Federated Learning}

\author{Bart Cox}
\email{b.a.cox@tudelft.nl}
\orcid{0000-0001-5209-6161}
\affiliation{%comment
  \institution{Delft University of Technology}
  \city{Delft}
  \state{}
  \country{Netherlands}
}
\renewcommand{\shortauthors}{Cox et al.}

\author{Abele M\u{a}lan}
\email{abele.malan@unine.ch}
\orcid{0009-0002-4493-7439}
\affiliation{%comment
  \institution{University of Neuch\^{a}tel}
  \city{Neuch\^{a}tel}
  \state{}
  \country{Switzerland}
}

\author{Lydia Y. Chen}
\email{lydiaychen@ieee.org}
\orcid{0000-0002-4228-6735}
\affiliation{%comment
  \institution{Delft University of Technology}
  \city{Delft}
  \state{}
  \country{Netherlands}
}
\author{J\'{e}r\'{e}mie Decouchant}
\email{j.decouchant@tudelft.nl}
\orcid{0000-0001-9143-3984}
\affiliation{%comment
  \institution{Delft University of Technology}
  \city{Delft}
  \state{}
  \country{Netherlands}
}

\begin{abstract}
Federated learning (FL) enables a set of
geographically distributed clients to collectively train a model
through a server. Classically, the training process is synchronous,
but can be made asynchronous to maintain its speed in
presence of slow clients and in heterogeneous networks. The
vast majority of Byzantine fault-tolerant FL systems however rely on
a synchronous training process. Our solution is one of the first Byzantine-resilient and asynchronous FL algorithms that does not require an auxiliary server dataset and is not delayed by stragglers, which are shortcomings of previous works. Intuitively, the server in our solution waits to receive a minimum number of updates from clients on its latest model to safely update it, and is later able to safely leverage the updates that late clients might send. We compare the performance of our solution with state-of-the-art algorithms on both image
and text datasets under gradient inversion, perturbation, and
backdoor attacks. Our results indicate that our solution trains
a model faster than previous synchronous FL solution, and
maintains a higher accuracy, up to 1.54x and up to 1.75x
for perturbation and gradient inversion attacks respectively, in the 
presence of Byzantine clients than previous asynchronous FL
solutions.

    %comment
    %comment
    %comment
    %comment
    %comment
    %comment
    %comment
    %comment
    %comment
    %comment
    %comment
    %comment
    %comment
    %comment
\end{abstract}


\settopmatter{printfolios=true}
\maketitle

\section{Introduction}

Federated Learning~\cite{mcmahan2017communication} aims to make good use of the vast amount of data distributed among a large number of users to train a model in a privacy-preserving way.
One of the key benefits of Federated Learning (FL) is to provide higher privacy guarantees than centralized machine learning~\cite{vandijk2020asynchronous}. By keeping the data on the local devices, FL can reduce the risk of data breaches or leaks that could compromise sensitive user information. FL also better follows data regulations such as the American Health Insurance Portability and Accountability Act (HIPAA)~\cite{hipaa2023} and the European General Data Protection Regulation (GDPR)~\cite{EuropeanParliament2016a}. Those concerns are particularly acute in settings where data is highly sensitive, such as healthcare or financial applications. 

Another advantage of FL is that it usually reduces the communication overhead associated with traditional centralized machine learning. Instead of sending large amounts of data to a central server, federated learning only requires small model updates to be exchanged between the devices and the server. This can help reduce network congestion and improve the speed and efficiency of the training process.

Asynchronous Federated Learning (AFL)~\cite{xie2019fedasync, wu2021safa, chai2020fedat} represents an innovative machine learning paradigm that revolutionizes the way global models are trained. Unlike its synchronous counterparts, \texttt{FedSGD} and \texttt{FedAvg}~\cite{mcmahan2017communication}, which necessitate all participating devices to engage in training simultaneously, AFL offers a unique level of flexibility. It permits devices to enter and exit the network at their own discretion, catering to scenarios where connections are unreliable or intermittent. Furthermore, it accommodates networks with heterogeneous computational resources and privacy concerns among their devices.
AFL's distinctive feature is its ability to facilitate asynchronous participation by nodes as devices can contribute to the model's evolution whenever they have available computational resources, without depending  on the progress of other nodes. This proves invaluable in situations where devices have limited computational capabilities, constrained bandwidth, or might go offline. 

In an AFL setup, each device maintains a local copy of the global model and independently conducts multiple rounds of training on its own dataset before exchanging updates with a central server or with other devices within the network. However, what truly sets AFL apart is the asynchronous nature of this update exchange. Unlike synchronous methods where updates are meticulously synchronized, AFL does not impose any specific order on the updates. As such, devices submit their updates independently and autonomously.
The central server or aggregator accumulates these updates in a manner that respects the devices' asynchrony. It subsequently employs these updates to construct a new global model. This freshly minted model is then redistributed to all participating devices to continue their individual training processes. This iterative cycle persists until the global model attains the desired level of accuracy or convergence.

However, preventing malicious attacks by clients in asynchronous FL~\cite{blanchard2017machine,cao2020fltrust,nguyenFLAMETamingBackdoors2022,mhamdiHiddenVulnerabilityDistributed2018} is more difficult than in synchronous FL because the server cannot wait for all client updates before aggregating them and updating the global model. 
Previous representative approaches, such as \texttt{Kardam}~\cite{damaskinos2018asynchronous}, \texttt{Zeno++}~\cite{xieZenoRobustFully2020} and \basgd{}~\cite{yangBASGDBufferedAsynchronous2021}, either make the strong assumption that the server has access to an auxiliary dataset, which is problematic from a privacy point of view, or has limited efficacy. In this work, we take inspiration from recent clustering methods that have been used to design Byzantine-resilient synchronous FL algorithms such as \texttt{Flame}~\cite{nguyenFLAMETamingBackdoors2022}. We design and evaluate \pname{}, a novel robust asynchronous FL algorithm that does not require an auxiliary server dataset and has higher performance than previous works in the attack-free case and under attack.      
As a summary, this work makes the following contributions. 

\begin{itemize}
\item  We identify the key building blocks of clustering-based Byzantine-resilient synchronous Federated Learning algorithms and decompose the \texttt{FLAME} algorithm to use them in an asynchronous manner.    
  
\item We design \pname{} the first asynchronous and Byzan-tine-resilient Federated Learning algorithm that does not require the server to hold an auxiliary dataset and is not delayed by stragglers. \pname{} requires at least $2f+1$ clients to tolerate $f$ Byzantine clients. Contrary to synchronous clustering-based algorithms, the first $2f+1$ client updates received for a global model are also used by \pname{} to trigger the computation of its next version, but late client updates computed by slow clients are also leveraged. We show that using a clustering-based approach in asynchronous settings requires paying attention to non-trivial but important design requirements.   
	
\item  We run an comprehensive evaluation of \pname{} in terms of accuracy and convergence speed with three datasets. We use two image datasets, namely \mnist{} and \cifar{}, and the \wikitext{} language dataset. We compare \pname{} to the \fasync{}, \texttt{Kardam}, and \basgd{} state-of-the-art baselines. Our results indicate that \pname{} converges faster and trains a model with higher accuracy under the gradient flipping, gradient inversion, and backdoor attacks than previous works.  
\end{itemize}

\noindent This paper is organized as follows. Section~\ref{sec:background} provides some background on Federated Learning and clustering-based Byzantine-resilient Federated Learning. Section~\ref{sec:models} details our system and threat models. Section~\ref{sec:intuition} provides a high-level overview of \pname{}, our asynchronous and Byzantine-resilient Federated Learning algorithm. Section~\ref{sec:system} discusses \pname{} in detail. Section~\ref{sec:evaluation} presents our performance evaluation. Finally, Section~\ref{sec:conclusion} concludes this paper. 

\section{Background}
\label{sec:background}

\subsection{Synchronous Federated Learning}

Federated Learning (FL) is a type of machine learning that enables multiple devices (e.g., laptops, mobile devices) to collaboratively train a global model without sharing their data with a central server. Instead, the data remains on the devices and client updates are aggregated by a server. This approach can help address some of the challenges associated with traditional centralized machine learning, such as data privacy, communication overhead, and scalability. 

FL involves the training of a machine learning model over distributed clients. In a federated system with $C$ clients, each client $c$ trains on its local dataset $\mathcal{D}_c=(x_c, y_c)$ and aims to minimize a loss function $F$ jointly with other clients.

In Federated Learning, in each round, the server selects a group of clients and sends them the latest global model. Upon receiving a global model $G^t$, a client $c$ trains it on its local dataset $D_c$ to compute $W_c^{t+1}$ as follows:  

\begin{equation*}
    W_c^{t+1} = G^{t} - \eta_c \frac{\partial F(G^t, D_c)}{\partial G^{t}},
\end{equation*}

where $F$ is the loss function of the classification, and $\eta_k$ is the learning rate. 
Afterwards, client $c$ sends its model update $W_c^{t+1}$ to the server. Upon receiving all client updates it expects in a given round, the server aggregates them to compute the next global model $G^{t+1}$ as

\begin{equation*}
    G^{t+1} = \sum_{c=1}^{C}{\frac{d_c}{d} W_{c}^{t}},
\end{equation*}

where, $d_c$ is the number of data points of client $c$ and $d$ is the total number of data points across all clients. The ratio $\frac{d_c}{d}$ is used to weight the participation of a client during aggregation. At the beginning of the following round, the server sends $G^{t+1}$ to the next subset of selected clients. \texttt{FedAvg} is however not robust under attack and sensitive to slow clients. 

\subsection{Asynchronous Federated Learning}

\begin{figure}[t]
	\centering
	%comment
	%comment
	\includegraphics[width=\columnwidth]{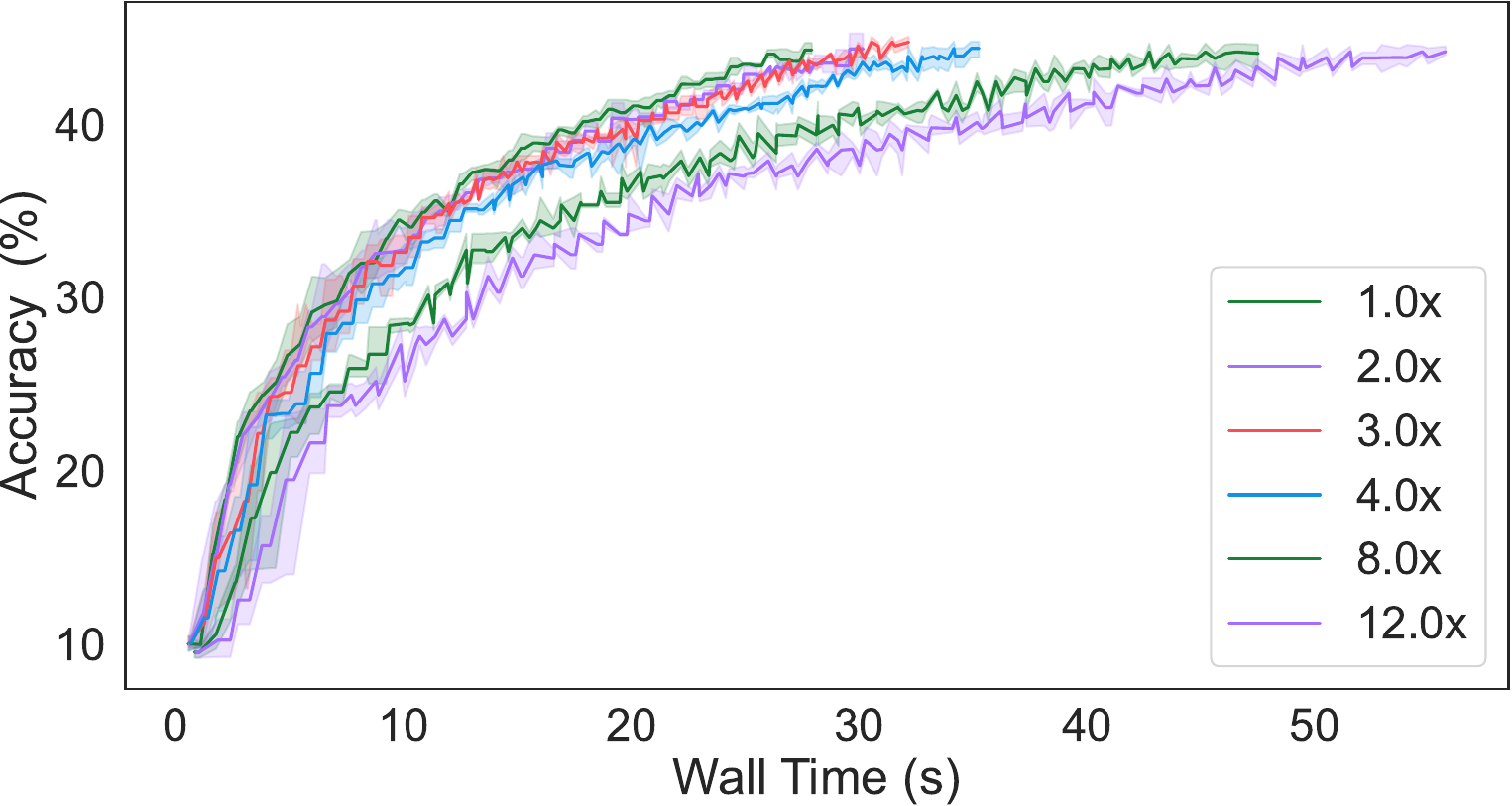}
	\caption{Wall clock time depending on the standard deviation of the client computing power distribution. Each line represent a distributed with modified standard deviation, e.g., the 2.0x line corresponds to a distribution whose standard deviation is twice as high as the one used with the 1.0x line. Higher client diversity slows down convergence. 
     }
 %comment
	\label{fig:need_for_afl}
\end{figure}

\begin{figure}[t]
	\centering
	%comment
	\includegraphics[width=\columnwidth]{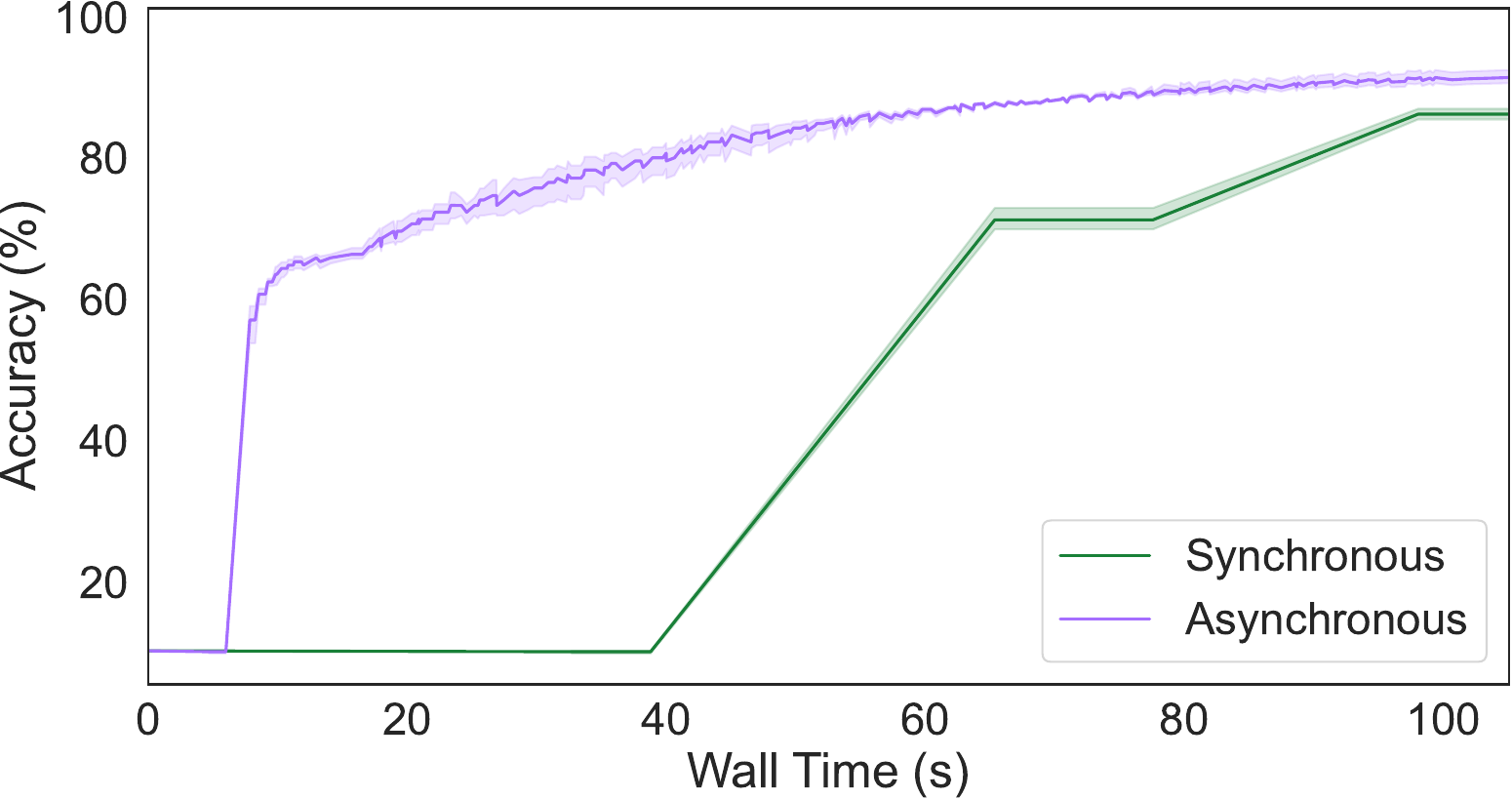}
	%comment
	\caption{Asynchronous FL requires less time to converge than synchronous FL. Dataset used is \mnist{}, number of clients is 40. 
		%comment
	}
	\label{fig:need_for_afl_1}
\end{figure}

In Figure~\ref{fig:need_for_afl} we illustrate the reduced performance of synchronous FL when clients are  heterogeneous. In this figure, the standard distribution of a set of $100$ heterogeneous clients was multiplied by 1, 2, 3, 4, 8 and 12 and the computing power of the clients resampled before running an experiment with the \cifar{} dataset, and the \texttt{FedAvg} algorithm. Consequently to those changes, the overall system slows down by $1.00$,  $1.08$, $1.15$,  $1.26$,  $1.70$, \text{and} $2.00$ times respectively.
 
One goal of asynchronous FL systems is to accelerate convergence in heterogeneous networks and under client heterogeneity. To do so, the server aggregates an update into a global model immediately after it receives it, and returns the new global model to the client. Figure~\ref{fig:need_for_afl_1} illustrates that asynchronous FL converges faster than  synchronous FL, when convergence speed is measured as elapsed time.  
For example, under the centralized asynchronous version of \texttt{FedAvg}, the server would update the global model as follows: 

\begin{equation*}
    G^{t+1} = G^{t} - s(\tau)\frac{d_c}{d}(W^{t-\tau}_{c}-W^{t-\tau+1}_{c})
\end{equation*}

\noindent where $G^{t}$ is the global model with timestamp $t$, $W^{c}_{t}$ is the local update of client $c$ with timestamp $t$, $\frac{d_c}{d}$ is the data proportion of the update, and $\tau$ is the difference between the ages of the current global model and the one the client update was computed with, and $s(\tau)$ is a staleness parameter that might dampen the effect of client $c$'s update if it relates to an old global model. 
Algorithm~\ref{alg:asyncsgd} provides the pseudocode of \fasync{}, the asynchronous version of the  \favg{} algorithm.

\begin{algorithm}[t]

    \caption{\fasync{}: Asynchronous Federated Learning}
    \label{alg:asyncsgd}
    \DontPrintSemicolon
    \SetInd{0.1em}{1em}
    \SetVlineSkip{0pt}
    \Parameter{$\tau$: staleness of the client model that the server receives}
    %comment
    %comment
    %comment

    \SetKwProg{Def}{~}{:}{}
    \Def{\textsc{Server}}{
    Global model $G^0 \gets$ random init \;
    $a \gets 0$ \tcp{age of the global server model} 
    send $G^0$ to all clients \;
    
    %comment
    \While {\textbf{not} (convergence achieved)} {
        upon receiving $W^c_{a-\tau} - W^c_{a-\tau+1}$  from client $c$, do $G^{a+1} \gets G^a - s(\tau) (G^{a} - W^{a-\tau+1}_c)$\;
        send $G^{a+1}$ to client $c$\;
        $a \gets a + 1$ \;
    }
    %comment
    %comment
    %comment
    }
    \;
    \Def{\textsc{Client $c \in [1,N]$}}{
        receive a global model $G^a$ from the server\;
        compute stochastic gradient $W^a_c$ based on $G^a$ and a random mini-batch of local training data\;
        send $W^{a+1}_c$ to the server\;
    }
\end{algorithm}

\subsection{Byzantine Federated Learning}

\begin{algorithm}[t]
    \DontPrintSemicolon
    \SetInd{0.1em}{1em}
    \SetVlineSkip{0pt}
    %comment
    %comment
    %comment
    %comment

    \SetKwProg{Def}{def}{:}{}

    \Def{\textsc{ClipBound}($W^1_t, \cdots, W^n_t$)}{ 
        $(e_1, \dots, e_n) \gets \textsc{EuclideanDistance}(G_t, \{W^1_t, \dots, W^n_t\})$ \;
        $S_t \gets \texttt{Median}(e_1, \dots, e_n)$ \;
        \Return $(e_1, \dots, e_n), S_t$
    }
    \;
    \Def{\textsc{Filtering}($W^1_t, \cdots, W^n_t$)}{    
        $(c_{11}, \dots, c_{nn}) \gets \textsc{CosineDistance}(W^1_t, \dots, W^n_t)$ \;
        $(b_1, \dots, b_L) \gets \textsc{Clustering}(c_{11}, \dots, c_{nn})$ \;
        \Return $\{ W^{b_l}_t : b_l \in  (b_1, \dots, b_L) \}$ \tcp{benign models}
    }
    \;
    \Def{\textsc{Aggregate}($G_t, (e_1, \dots, e_n), S_t, (W^{b_1}_t, \cdots, W^{b_L}_t)$)}{ 
        \For{each client $l \in [1, L]$} {
            $W^{b_l}_t {\gets} G_t {+} (W^{b_l}_t {-} G_t) {\cdot} \mathrm{min}(1, \frac{S_t}{e_{b_l}})$ \tcp*[h]{clipping}
        } 
        $G_{t+1} \gets \sum_{l=1}^{L} W^{b_l}_t / L$ \tcp{aggregation}
        $\sigma \gets \lambda \cdot S_t$ where $\lambda = \frac{1}{\epsilon} \cdot \sqrt{2 ln(\frac{1.25}{\delta})}$ \;
        $G_{t+1} \gets G_{t+1} + N(0, \sigma^2)$ \tcp{noising}
        \Return $G_{t+1}$
    }
    \;
    \Def{main()}{
        $G_0 \gets$ random init \tcp{global server model}
        \For{\text{each training iteration} $t \in [0,T-1]$} {
            send $G_t$ to all clients \;
            \For{each client $i \in [1,n]$} {
                $W^i_t \gets ClientUpdate(G_t)$
             }
             $(e_1, \dots, e_n), S_t = \textsc{ClipBound}(W^1_t, \cdots, W^n_t)$ \;
             $correct = \textsc{Filtering}(W^1_t, \cdots, W^n_t)$ \;
             $G_{t{+}1} = \textsc{Aggregate}(G_t, (e_1, \dots, e_n), S_t, correct)$ 
        }
    }

    %comment
    %comment
    %comment

    %comment
    %comment
    %comment
    %comment
    %comment
    %comment
    %comment
    %comment
    %comment
    %comment
    %comment
    %comment
    %comment
    %comment
    %comment
    %comment
    %comment
    %comment
    %comment
    %comment
    %comment
    %comment
    %comment
    %comment
    %comment
    %comment
    %comment
    %comment
    %comment
    %comment
	\caption{Pseudocode of the \texttt{FLAME}~\cite{nguyenFLAMETamingBackdoors2022} synchronous FL Byzantine-resilient algorithm. The code has been refactored around three functions, \textsc{ClipBound}, \textsc{Filtering} and \textsc{Aggregate}, that \pname, our asynchronous Byzantine-resilient FL algorithm, leverages.} 
	\label{alg:flame}
\end{algorithm}

Byzantine FL aims to protect the integrity and accuracy of the training process in the presence of clients that behave maliciously or arbitrarily. In a Byzantine fault model, nodes can crash or provide incorrect information in unpredictable ways, which can compromise the accuracy of the model and system security. Byzantine learning seeks to develop algorithms and protocols that are resilient to these types of faults, so that the system can continue to operate effectively and accurately even in the presence of adversarial behavior. 

Current defenses based on clustering~\cite{damaskinos2018asynchronous}, aim to detect potentially malicious model updates by organizing them into two clusters. The smaller cluster is consistently labeled as malicious, and its updates are subsequently eliminated. However, in the case that there are no Byzantine clients present, a subset of the benign updates will be filtered out. Instead, more recent clustering approaches find the biggest cluster of updates, which ensures that the majority set of the most similar updates is used during the aggregation, independently of the possible presence of Byzantine updates. Using the angular distance between vectors, the biggest cluster can be determined as benign updates if updates from $2f+1$ clients are used.

As \pname{} relies on clustering and filtering algorithms, we now introduce as an example of the key ideas behind \texttt{FLAME}~\cite{nguyenFLAMETamingBackdoors2022}, which is a state-of-the-art Byzantine-resilient synchronous FL algorithm that relies on client model clustering. We refer the reader to Section~\ref{sec:sota} for a more exhaustive discussion of Byzantine-resilient FL algorithms. 
Algorithm~\ref{alg:flame} presents \texttt{FLAME}'s pseudocode. Compared to its original monolithic version, this code contains  functions that encapsulate different key roles.  At the beginning of a round, the server sends the current global model $G_t$ to all clients. Each client then trains the global model on its dataset and sends its update $ClientUpdate(G_t)$ to the server. Upon receiving all client updates, the server first computes and clips all gradients, i.e., reduces their amplitude, using the \textsc{ClipBound} function. This function eventually returns all clipped gradients and their median $S_t$. The \textsc{Filtering} function then clusters all clipped gradients using the HDBSCAN clustering algorithm~\cite{campello2013density} and returns the biggest cluster of gradients, which contains the correct models when a minority of the clients are Byzantine. Finally, the \textsc{Aggregate} function aggregates all clipped and correct gradients, and uses them to update the global model and obtain its new version $G_{t+1}$. 

\texttt{FLAME} is synchronous since the server waits for all client updates before updating the global model. Our protocol, \pname{}, also makes use of the \textsc{ClipBound}, \textsc{Filtering} and \textsc{Aggregate} functions. However, when the number of clients is sufficient, \pname{} does not wait for all of them to return their model updates and therefore allows fast and slow clients to train different global model versions, which results in increased training speed. To the best of our knowledge, \pname{} is the first Byzantine-resilient asynchronous FL algorithm that relies on client model clustering.  

\section{System and Threat Models}
\label{sec:models}

\subsection{System Model}

The federated learning system is composed of one server and $N$ clients. Each client has a local dataset that it uses to train a version of the global model and generate a client update that it sends to the server. We consider the horizontal FL use case: clients hold different data instances that share the same features. However, client datasets might be Independent and Identically Distributed (IID) or not.  
The server is responsible for maintaining the global model. 

\subsection{Threat Model}

We assume that up to $f$ clients in the system can exhibit Byzantine behavior, i.e., they can behave arbitrarily or maliciously. Similar to previous works~\cite{damaskinos2018asynchronous,yangBASGDBufferedAsynchronous2021,xieZenoRobustFully2020, blanchardMachineLearningAdversaries2017}, the server is assumed to be correct. 
We assume the standard asynchronous parameter model~\cite{abadi2016tensorflow}. We assume that at most $f < \floor{N/2}$ clients are Byzantine.
In particular, Byzantine clients can launch targeted or untargeted attack on the FL system. Byzantine nodes can operate as singular nodes or collude. Byzantine nodes have unbounded computational power and arbitrarily fast communication.

A Byzantine client proposes a gradient that can deviate arbitrarily from gradients proposed by correct clients. 
Furthermore, the attacker may possess different levels of knowledge about the Federated Learning (FL) system~\cite{cao2020fltrust,DBLP:conf/uss/FangCJG20}. This knowledge can be categorized into partial knowledge and full knowledge. In the partial-knowledge setting, the attacker is aware of the local training data and model updates on the malicious clients. On the other hand, in the full-knowledge scenario, the attacker has comprehensive knowledge of the entire FL system. This includes knowing the local training data and model updates on all clients, as well as understanding the FL method and its parameter settings. It is emphasized that an attacker with full knowledge is considerably more potent than one with only partial knowledge~\cite{DBLP:conf/uss/FangCJG20}. 

The sequence of interaction between the clients and the server is particularly important in asynchronous learning. We assume that a Byzantine node has full knowledge of the times at which all nodes will send their model update to the server. This makes it possible for Byzantine nodes to "inject" their update just before or just after any regular client's update.

\section{\pname{} in a Nutshell}
\label{sec:intuition}

\subsection{Clustering}

In asynchronous Byzantine Learning, the interaction between client and server is one-to-one: the server gets an update on a model from only one client. Client updates are always used to update the global model. However, this leaves the server with little information whether to accept or reject a client update, and therefore make the system more vulnerable to Byzantine attacks. 

On the opposite side of the spectrum, in synchronous systems at the beginning of a round, the server asks clients to all train the same model on their local data. When a client sends its weights to the server, the server only aggregates a new version of the model if it has updates from enough distinct clients.
In order to aggregate client updates in a robust manner one can wait for $2f+1$ updates from $2f+1$ different clients and use a clustering method (Alg.~\ref{alg:flame}) to identify and only use the benign updates. In asynchronous settings however, this synchronous FL approach would then lead the server to discard the slow updates that would arrive after the $2f+1$ first ones, which would waste the resources slow clients dedicated to compute their updates and would additionally delay the global model convergence.  

\pname{} applies a clustering approach on the $2f+1$ first client updates it receives to filter out Byzantine client updates and update the global model, however, it also makes use of slow client updates. 

\subsection{Slow Clients and Liveness}

When adapting Byzantine clustering for asynchronous FL, the effect of slow clients has to be leveraged and kept under control. A slow client can either produce a stale model update or be a Byzantine client. 
When a client responds to the server, it checks if the client is responding to the current round or to an older round. Updates from old rounds are validated before using it for a new model update. The validation is done using the computed cluster from the faster $2f+1$ clients that already arrived.
By re-clustering with the $2f+1$ faster clients and the additional update from slow clients, we can investigate if the clusters changed. 
Validated slow updates can then be used to update the global model the next time it is updated using $2f+1$ updates that are received on the latest global model. In the meantime, the server also returns the latest global model to slow clients that send a late update so that they always remain active and so that $2f+1$ updates on the latest global model are always eventually collected. 

\subsection{Number of Clients}

\pname{} requires at least $2f+1$ clients to tolerate $f$ Byzantine clients. In particular, \pname{} would behave as a synchronous algorithm when there are exactly $2f+1$ clients in the system.  However, contrary to synchronous solutions, \pname{}'s convergence speed increase (measured as elapsed time) compared to robust synchronous FL solutions is more significant when the system contains a larger number of clients.     
This characteristic of \pname{} is however not a limitation compared to previous works, which even though they claim to tolerate a large proportion of faulty clients presents performance evaluation where the proportion of faulty nodes most often ranges from 10\% to 40\% of the clients~\cite{yangBASGDBufferedAsynchronous2021, nguyenFLAMETamingBackdoors2022, blanchardMachineLearningAdversaries2017,xiaByzantineTolerantAlgorithms2023a}. 

\section{\pname{} Details}
\label{sec:system}

\begin{algorithm*}
    \DontPrintSemicolon
    \SetInd{0.1em}{1em}
    \SetVlineSkip{0pt}
    %comment
    \Parameter{${sf}_\alpha(a, i)$: staleness function (e.g., $\alpha/(a-i)$)}
    \KwParameterXX{$\eta$: learning rate} %comment
    \KwParameterXX{$C$: set of all clients}
    \KwParameterXX{$K$: number of previous models for which updates are processed} 

    \SetKwProg{Def}{def}{:}{}

    \Def{Main()}{
        %comment
        %comment
        $G_0 \gets$ random model \tcp{$G_i$ is the i-th global server model}
        $a \gets 0$ \tcp{age of the global server model} 
        $pending \gets \emptyset$ \tcp{map (age: set of weights) for updates   waiting to be included in the global model}
        $processed \gets \emptyset$ \tcp{map (age: set of weights) for updates that have been included in the global model}
        $age\_client\_models \gets [0, \cdots, 0]$ \tcp{age of the global model that clients are training} 
        $rcvd\_model \gets \emptyset$ \tcp{only one update per client per model age}
        $idle\_clients \gets \emptyset$ \tcp{set of fast clients waiting for next global model}
        $S \gets \emptyset$ \tcp{Median of 2f+1 fastest updates that triggered a global model update}
        \BlankLine
        \BlankLine
        send $W^s_0$ to all clients in $C$ \label{line:model_to_clients} \;

        \While {\textbf{not} (convergence achieved) \textbf{and} upon receiving $W^c_t$  from client $c$} {
        %comment
            $t = age\_client\_models[c]$ \;
            \uIf{$(c, t) \in rcvd\_model$\label{line:check_rcvd_model}} {
                \textbf{continue} \;
            }
            $rcvd\_model \gets rcvd\_model \cup \{ (c,t) \}$ \;
            \uIf{$(t = a)$ \label{line:check_age}} {
                $pending[a] \gets pending[a]\, \cup\, \{W^c_t\}$ \;
                \uIf(\tcp*[h]{between $2f{+}1$ and $N{-}f$}){$|pending[a]| \geq 2f{+}1$\label{line:check_pending}} {
                    \For{$i \in [a{-}K{+}1, a{-}1]$} {
                        $(e_1, \cdots, e_{n_i}), - \gets \textsc{Clipbound}(processed[i] \cup pending[i])$ \label{line:k_clipbound} \;
                        $(W^{b_1}_i,\cdots,W^{b_l}_i) \gets \textsc{Filtering}(processed[i] \cup pending[i])$ \label{line:k_filtering} \;
                        %comment
                        $\overline{W_{i}} \gets \textsc{Aggregate}\left(G_i, (e_1, \cdots, e_{n_i}), S[a], (W^{b_1}_i,\cdots,W^{b_l}_i)\, \cap\, pending[i]\right)$ \label{line:k_aggr} \tcp{reuse past clipping bound}
                    }
                    $(e_1, \dots, e_{n_a}), S[a] \gets \textsc{ClipBound}(pending[a])$ \label{line:g_clipbound} \tcp{clip. bound of the first $2f+1$ updates}
                	$(W^{c_1}_a,\cdots,W^{c_k}_a) \gets \textsc{Filtering}(pending[a])$ \label{line:g_filtering} \;
                    $\overline{W_a} \gets \textsc{Aggregate}\left(G_a, (e_1, \dots, e_{n_a}), S[a], (W^{c_1}_a,\cdots,W^{c_k}_a)  \right)$ \label{line:g_aggr}\; 
                    $G_{a{+}1} \gets \overline{W_a} - \sum_{i=a{-}K{+}1}^{a{-}1} {sf}_\alpha(a,i) \cdot \frac{|pending[i]|}{|C|} \cdot \eta (\overline{W_i}-G_i)$ \label{line:updated_model} \;
                    %comment
                    $a \gets a {+} 1$\; 	
                    \For{$i \in [a{-}K{+}1, a]$} {
                        $processed[i] \gets processed[i] \cup pending[i]$ \;
                        $pending[i] \gets \emptyset$  
                    }
                    \For{$d \in idle\_clients \cup \{c\}$} {
                        $age\_client\_models[d] = a{+}1$ \;
                        send $G_{a{+}1}$ to client $d$ \;
                    }
                 	  $idle\_clients \gets \emptyset$ \;
                    $processed.delete(a{-}K{+}1)$ \tcp{delete key and associated data} 
                    $pending.delete(a{-}K{+}1)$ \; 
                    $S.delete(a{-}K{+}1)$ \;
                    $a \gets a {+} 1$ 
                }
            	\Else{
                    $idle\_clients \gets idle\_clients \cup \{c\}$ \tcp{client $c$ will wait for next global model}
            	}
            }
            \Else{\label{line:delayed_update}
                \uIf(\tcp*[h]{Delayed update for one of the prev. $K-1$ rounds}){$(t \in [a-K+1, a-1])$} {
                    $pending[t] \gets pending[t] \cup \{W^c_t\}$ \label{line:fast_client_to_pending} \;
                }
                %comment
                $age\_client\_models[d] = a$ \;
                send $G_{a}$ to client $c$ \label{line:send_current_model}\;
            }
        }
    }
	\caption{\pname{}}
	\label{alg:pess_server_3}
\end{algorithm*}

Algorithm~\ref{alg:pess_server_3} details \pname{}'s pseudocode, which employs the ideas we have discussed in the previous Section. The clients in \pname{} behave exactly like they would in \fasync{}. However, the server is doing additional work to filter client updates and aggregate them. In the following, we focus on the code that the server executes. 

\subsection{Variables and Main Loop}
 The server maintains several variables. It first keeps track of the age $a$ of its global model, which is the number of times the global model has been updated, each time based on $2f+1$ client updates computed on the previous global model. The server also accumulates the client updates received and is waiting to process in a map $pending$, and remembers the client updates it has already processed in another map $processed$. The age of the global model the clients are training on is stored in an array $age\_client\_models$, and a map $rcvd\_model$ remembers whether a client has sent a client update for a given global model age. Finally, the $idle\_clients$ set stores the identity of fast clients that return updates computed on the latest global model and that are waiting for the next one. 

The algorithm starts by requesting all clients to train the initial randomly selected global model $G_0$ (l.~\ref{line:model_to_clients}).
When a client responds to the server, the server checks that the update is the first update that the client sends for a given global model (l.~\ref{line:check_rcvd_model}). If this is not the case, then the update comes from a Byzantine client and is ignored. 
If this verification is successful, the server then determines whether the client update  received corresponds to the latest global model (in which case it continues with l.~\ref{line:check_age}) or to an older global model (in which case it continues with l.~\ref{line:delayed_update}).

\subsection{Processing Late Client Updates}

Upon receiving a late model update, the server verifies whether it is computed on one of the latest $K$ global models. If so, it inserts the client update in the $pending$ map of updates. The server then sends the latest global model $G_a$ to the late client (l.~\ref{line:send_current_model}) to allow it to possibly catch up with faster nodes and leverage its dataset. Note that the server could detect whether a client is becoming too late and keeps working on a global model that is too old. In that case, the server can directly send the latest global model to such a client. 

\subsection{Processing Fast Client Updates}
Upon receiving a client update computed using the latest global model, the server adds the weights to the $pending$ map.  If the size of the set of updates that are used to compute on the latest global model,  $pending[a]$, is lower than $2f+1$, then update filtering cannot yet take place and the server  lets the client wait for the next global model to be computed. To do so, client $c$ is added to the $idle\_clients$ set (l.~\ref{line:fast_client_to_pending}). 

If the size of $pending[a]$ is equal or larger than $2f+1$ (l.~\ref{line:check_pending}), then the next global model can be computed. All late updates that have been received since the last global model computation, which are stored in $pending[a-K+1]$, ..., $pending[a-1]$, are then clipped, clustered with their previously received counterparts and filtered (ll.~\ref{line:k_clipbound}-\ref{line:k_filtering}). The benign late updates that correspond to each previous global model are then aggregated (l.~\ref{line:k_aggr}) and stored in a variable $\overline{W}_i$. In case a correct fast client update that was considered faulty at the time a global model was updated is now evaluated to be correct, which is susceptible to happen at the beginning of the training process, it can then also be used to update the current global model (not shown in the pseudocode for simplicity). It is important to notice that the clipping bound that was used for the first $2f+1$ updates of a global model need to be reused for late updates in this aggregation step (l.~\ref{line:g_filtering}), otherwise late updates would be given too much weight. The $2f+1$ fast client updates computed on the latest global model are also clipped, filtered and aggregated (ll.~\ref{line:g_clipbound}-\ref{line:g_aggr}). 

Finally, the new global model is computed using both the fast and the slow client updates that have been received since the last global model was computed (l.~\ref{line:updated_model}).   
More precisely, the new global model $G_{a{+}1}$ is computed as 

\begin{equation*}
    G_{a{+}1} \gets \overline{W_a} - \sum_{i=a{-}K{+}1}^{a{-}1} {sf}_\alpha(a,i) \cdot \frac{|pending[i]|}{|C|} \cdot \eta (\overline{W_i}-G_i)
\end{equation*}

This formula contains several interesting details. First, ${sf}_\alpha(a,i)$ is a staleness function that dampens the effect of late updates on the model. We use ${sf}_\alpha(a,i) = \alpha/(a-i)$ in our experiments and set parameter $\alpha$ to 1. Second, a term $\frac{|pending[i]|}{|C|}$ also allows late updates to have a proportional impact of the global model.  
Once the new global model has been computed, the server sends it to all fast clients that are waiting for it (l. 32). The server also updates its variables $processed$ and $pending$ (ll. 30 and 31), and garbage collect the entries that correspond to a past global model that will not be used anymore (ll. 36 -- 38). 

\section{Performance Evaluation}
\label{sec:evaluation}
\subsection{Experimental Setup}
\begin{figure}[ht]
	\centering
	%comment
	\includegraphics[width=\columnwidth]{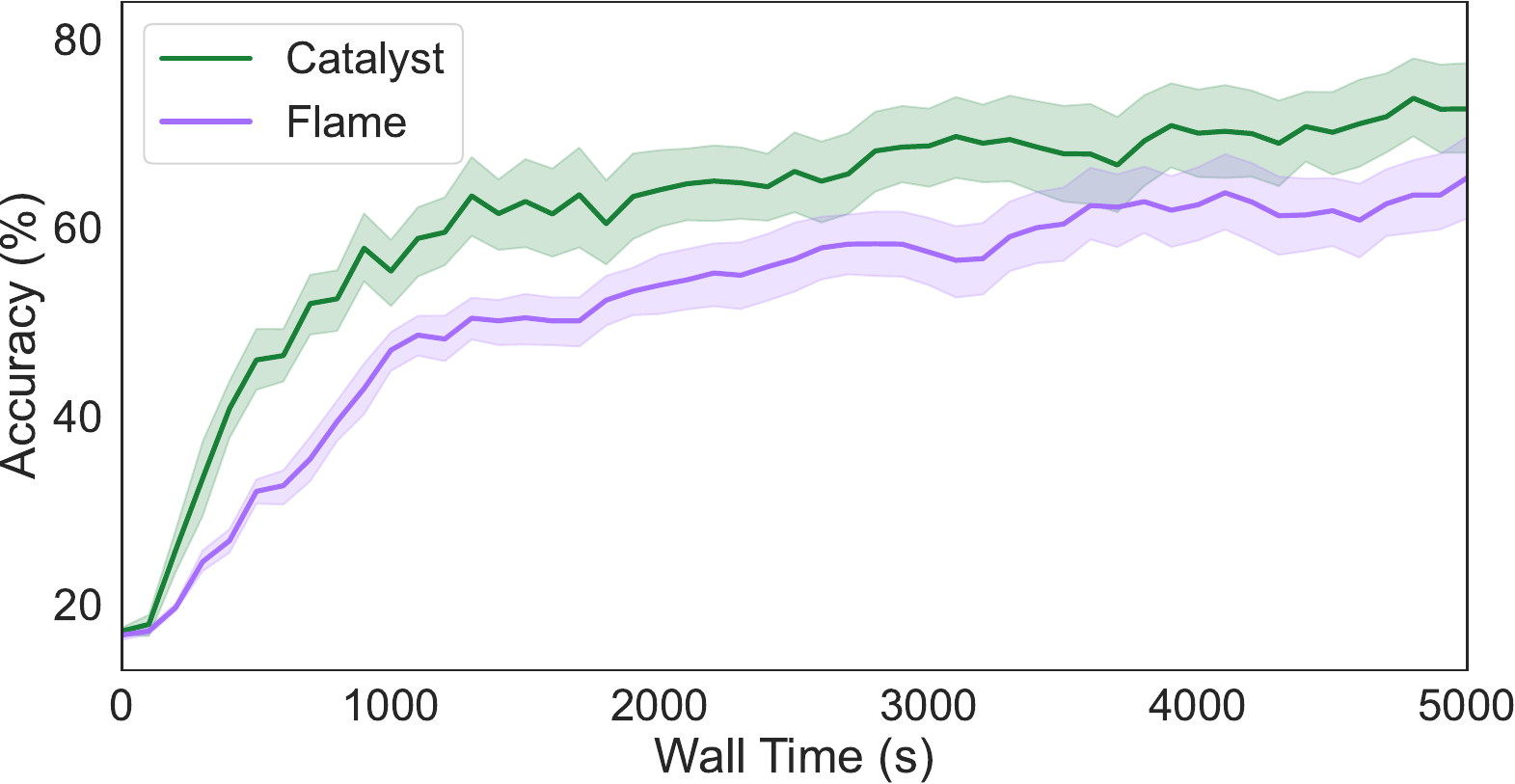}
	\caption{\pname{} and Flame with asynchronous clients.}
	\label{fig:catalyst_vs_flame_baseline}
\end{figure}
We use the PyTorch deep learning framework~\cite{PyTorch} and use the source code provided by Bagdasaryan et al.~\cite{bagdasaryan2020backdoor}, Blanchard et al.~\cite{blanchard2017machine} and Fang et al.~\cite{DBLP:conf/uss/FangCJG20} to implement the attacks.
We reimplemented existing defenses to compare them with \pname{}.

\subsubsection{Datasets and Learning Configurations}

We evaluate \pname{} in two typical application scenarios: image classification~\cite{HeZRS16_residual, OhKY22Moon, LiHS21Contrastive} and word prediction \cite{LinHM0D18,mcmahan2017communication} as summarized in Table~\ref{tab:time-perf} and Table~\ref{tab:atk_backdoor}.

For the neural network models, we use a CNN model with two convolutional layers and two fully connected layers to perform \mnist{}~\cite{lecun1998gradient} classification. For \cifar{}~\cite{krizhevsky2009learning}, a CNN model with three convolutional layers and two fully connected layers is used. For the text dataset \wikitext{}~\cite{merity2016pointer}, we use the next character long short-term memory (LSTM) neural network model, designed for character-level text generation tasks. The language model consists of an embedding layer that is used to capture the semantic and syntactic properties of characters, an LSTM layer that captures dependencies between characters and generates coherent text, and a fully connected layer that transforms the LSTM's hidden states into a probability distribution over all possible characters in the vocabulary.

\subsubsection{Baselines}

We selected baselines that contain state-of-the-art algorithms for both the fault-free and Byzantine adversarial models. \fasync{}~\cite{xie2019fedasync} is a performant asynchronous method that assumes that no adversaries are present. \texttt{Kardam}~\cite{damaskinos2018asynchronous} attempts to filter information received from deviant clients. We set the strength of its filtering parameter $\gamma = 0.1$, as in other works~\cite{yangBASGDBufferedAsynchronous2021}. \basgd{} employs multiple buffers to aggregate received models, computing the average of updates within each buffer, before further applying a function over the resulting average to yield the actual updated model. We use the median as the aggregation function, as it performed best in the original author's experiments~\cite{yangBASGDBufferedAsynchronous2021}, and set the number of buffers $B = 2f + 1$, per their recommendation. For \pname{} we set $K = 5$, the number of client updates necessary to trigger a global model update aggregation to $\max\{2, 2f + 1\}$, and the delay impact $d = 1$.

\subsubsection{Attacks}

We use the following data poisoning attacks in our experiments. 
\begin{itemize}
\item \textbf{Random Perturbation (RP) attack}~\cite{blanchard2017machine}: Each model update from malicious clients is drawn from a zero-mean Gaussian distribution (we set the standard deviation to $\norm{\sigma_{atk}}^2$, where $\sigma_{atk}$ is the attack parameter and $g$ the computed gradient by the client).
\item \textbf{Gradient Inversion (GI) attack} \cite{DBLP:conf/uss/FangCJG20}: A malicious client computes a model update based on its local training data and then scales it by a negative constant (-10 in our experiments) before sending it to the server.
\item  \textbf{Backdoor (BD) attack} \cite{DBLP:badnets, bagdasaryan2020backdoor}: BD attack is a targeted poisoning attack. We use the same strategy as in \cite{DBLP:badnets} to embed the trigger in \mnist{}. %comment
\end{itemize}

\subsubsection{Evaluation Metrics}
We measure the accuracy of the trained models for image classification, and its perplexity for the language models.  For classification on \mnist{} and \cifar{}, higher accuracy denotes better performance, while for text generation performance it is inversely proportional to the measured perplexity.
In the evaluation of backdoor attack and defense techniques, we consider a specific set of metrics. The Backdoor Accuracy ($BA$) measures the accuracy of the model in the backdoor task. It represents the fraction of the trigger set for which the model provides incorrect outputs as chosen by the adversary. The adversary seeks to maximize BA, while an effective defense aims to prevent the adversary from achieving this.
The Main Task Accuracy ($MA$) assesses the accuracy of a model in its main (benign) task. It indicates the fraction of benign inputs for which the system provides correct predictions. The adversary's goal is to minimize the impact on MA to reduce the likelihood of being detected. A robust defense system should not adversely affect MA.

\subsection{Wall Time Efficiency}

We start by examining the performance of the models resulting from different algorithms after a point in time close to the convergence of the fastest method. We set the learning rate of all clients and all servers that use one to the same value for each dataset: $0.05$ for \mnist{}, $0.25$ for \cifar{}, and $1.0$ for \wikitext{}. Furthermore, samples within clients are non-IID for the two classification tasks, while sampling is uniform for the text generation task. All clients' compute times (in seconds) fit from a normal distribution with a mean of 100 and a standard deviation of 20.

Table~\ref{tab:time-perf} shows the results we obtained for our three tasks and the examined baselines, along with \pname{}. 
The left-hand side of the table shows the performance for the no-attack ($f = 0$) case. Here, thanks to its insensitivity to learning rate, \texttt{Kardam} performs best over the image datasets, while on the other end of the spectrum, \fasync{} often struggles already. When switching to text data, \basgd{} and \pname{} perform very similarly, besting the others. Overall, \pname{} provides consistently high performance across all tested datasets before and after introducing attackers.

The middle column of Table~\ref{tab:time-perf} tackles the random perturbation attack, where both \fasync{} and \texttt{Kardam} show weaknesses, often not running long enough to reach the target wall time. \basgd{} performs well, and \pname{} does even better. \pname{} has 5 or more additional percentage points of accuracy over \basgd{} in both \mnist{} and \cifar{}, with the lead in \wikitext{} being even more substantial at over 600 points. \pname{} still achieves scores similar to its own when no attack exists. The tested instance of the attack has strength $\sigma{=}0.1$.

Finally, the right-hand side of the table showcases the gradient inversion attack. Neither \fasync{} nor \texttt{Kardam}, which has some protection against attacks, complete a full evaluation cycle. \basgd{} and \pname{} retain their relative positions, but the latter builds up an even greater gap to the former within image data.

\begin{table*}[htbp]
    \centering
    \tabcolsep=0.1cm
	\rowcolors{2}{gray!10}{white}
    %comment
    %comment
    %comment
    %comment
    %comment
    %comment
    %comment
    %comment
    %comment
    %comment
    %comment
    %comment
    %comment
    \begin{tabular}{llllllllll}
        \toprule
        \multirow{2}{*}{Algorithm} & \multicolumn{3}{c}{No Attack} & \multicolumn{3}{c}{Random Perturbation Attack} & \multicolumn{3}{c}{Gradient Inversion Attack}\\
        \cmidrule(lr){2-4} \cmidrule(lr){5-7} \cmidrule(lr){8-10}
& \multicolumn{1}{c}{\mnist{}} & \multicolumn{1}{c}{\cifar{}} & \multicolumn{1}{c}{\wikitext{}} & \multicolumn{1}{c}{\mnist{}} & \multicolumn{1}{c}{\cifar{}} & \multicolumn{1}{c}{\wikitext{}} & \multicolumn{1}{c}{\mnist{}} & \multicolumn{1}{c}{\cifar{}} & \multicolumn{1}{c}{\wikitext{}} \\
        \midrule
        \fasync{}        & 62.1$\pm$8.6 & 50.4$\pm$0.7 & 1507$\pm$72 & 25.5$\pm$4.3   & 10.0$\pm$0.0* & 18335$\pm$3751* & 10.0$\pm$0.2   & 10.0$\pm$0.0* & NA \\
        \texttt{Kardam}          & 92.0$\pm$0.5 & 64.8$\pm$2.3 & 1341$\pm$32 & 48.0$\pm$12.9* & 10.0$\pm$0.0* & 18508$\pm$3767* & 36.0$\pm$19.3* & 10.1$\pm$0.1* & NA \\
        \basgd{}           & 66.9$\pm$1.2 & 42.8$\pm$1.2 & 1062$\pm$41 & 86.8$\pm$3.3   & 55.5$\pm$0.3  & 1854$\pm$471    & 82.0$\pm$1.4   & 39.3$\pm$4.4  & 1878$\pm$128 \\
        \textit{\pname{}} & 89.4$\pm$2.4 & 64.6$\pm$1.9 & 1060$\pm$72 & 92.0$\pm$0.4   & 62.2$\pm$1.8  & 1203$\pm$61     & 92.0$\pm$0.4   & 68.7$\pm$1.6  & 1509$\pm$466 \\
        \bottomrule
    \end{tabular}
    \caption{Performance (accuracy for \mnist{} \& \cifar{} -- higher is better, perplexity for \wikitext{} -- lower is better) of each algorithm after some number of seconds (750 for \mnist{}, 7000 for \cifar{}, 600 for \wikitext{}). Each experiment is repeated three times. The values are the average time and the standard deviation. A * denotes that the algorithm did not run up to the selected time threshold. NA denotes the lack of any valid result. There are 40 clients including 10 faulty ones.}%comment
    \label{tab:time-perf}
\end{table*}

\subsection{Importance of slower clients}
While Flame show a good performance in synchronous FL, it have difficulties in asynchronous scenarios. This is due to the fact that FLAME assumes it will see all clients that participate in a round when it filters the Byzantine nodes. We adapted Flame to work in an asynchronous scenario.  Both algorithms wait for at least $2f+1$ clients before aggregation can happen. In scenarios where the total number of nodes in the system is bigger than $2f+1$, slow clients are at risk of being ignored. This can be problematic when important data points are located at the slower clients. Figure~\ref{fig:catalyst_vs_flame_baseline} shows that Flame performs worse than \pname{} in scenarios where $f$ is relatively small compared to the total number of clients.

\subsection{Resilience to Backdoor Attacks}

Results on the tested backdoor attacks for images once again show that \pname{}, our proposed algorithm, performs best, having the lowest BA while maintaining a similar MA with or without an attack. In terms of minimizing BA, the buffered \basgd{} represents a marked improvement against the defenseless \fasync{}, while \texttt{Kardam}'s filtering approach proves almost three times as effective. With a BA $<1$, our approach is an order of magnitude better than \texttt{Kardam}.

For the other backdoor-specific metric, MA, all examined methods maintain similar accuracy after becoming attack victims. Interestingly, \fasync{}'s defenseless design also makes it slightly more susceptible to performance degradation when dealing with images without a backdoor.

\begin{table}[htbp]
    \centering
	\rowcolors{2}{gray!10}{white}
    \begin{tabular}{l|ll}
        %comment
         & \multicolumn{2}{c}{\mnist{}}\\
        \cmidrule(lr){2-3}
        Algorithm & {BA} & {MA} \\
        \midrule
        No Byzantine & \textemdash & 97.8 \\
        No defense & 99.1 & 97.4 \\
        \midrule
        \fasync{} & 98.5 &     93.89 \\
        \texttt{Kardam} & 12.7 &     97.75 \\
        \basgd{} & 34.4 &     95.94 \\
        \pname{} (This work) & 0.5 &     96.95 \\
        %comment
        %comment
    \end{tabular}
    \caption{Backdoor Accuracy (\textit{BA}) and Main Task Accuracy (\textit{MA}) on \mnist{} dataset. All values are percentages. }
    \label{tab:atk_backdoor}
\end{table}

\subsection{Long-term Performance Trends}

To provide a more comprehensive comparison with the baselines over the different parts of a training cycle, we plot the accuracy/perplexity of resulting models concerning the wall time. The settings for (Byzantine) clients and the servers are the same as for Table~\ref{tab:time-perf}. Due to its greater disruption capabilities, we focus on the no-attack case as a preliminary check and gradient inversion.

\begin{figure*}[htp]
	\centering
	\begin{subfigure}[t]{0.45\textwidth} %comment
		\centering
		\includegraphics[width=\columnwidth]{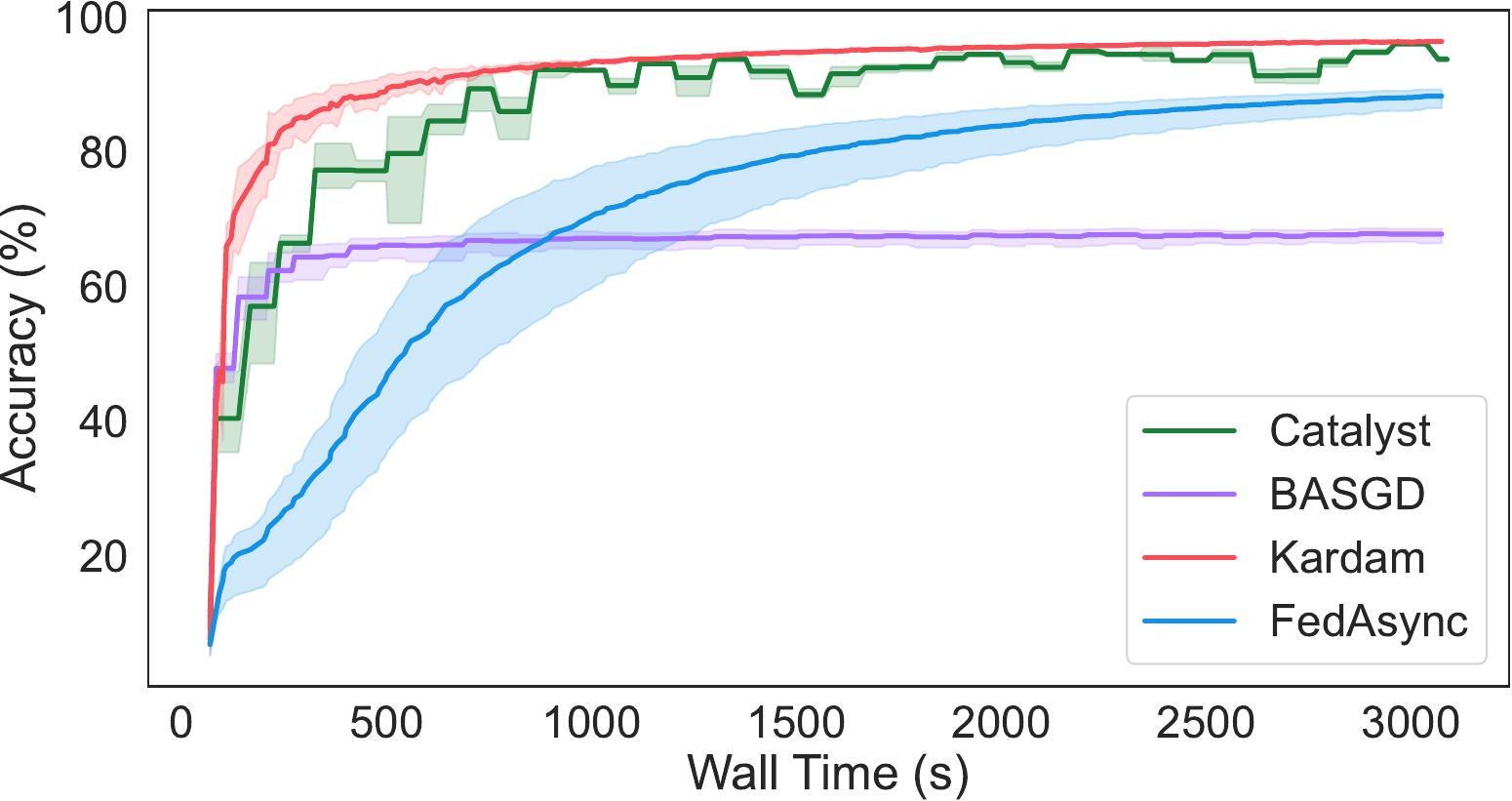}
		\caption{Accuracy with \mnist{} with no attacks}
		\label{fig:acc_mnist_no_atk}
	\end{subfigure}
	\hfill
	\begin{subfigure}[t]{0.45\textwidth} %comment
		\centering
		\includegraphics[width=\columnwidth]{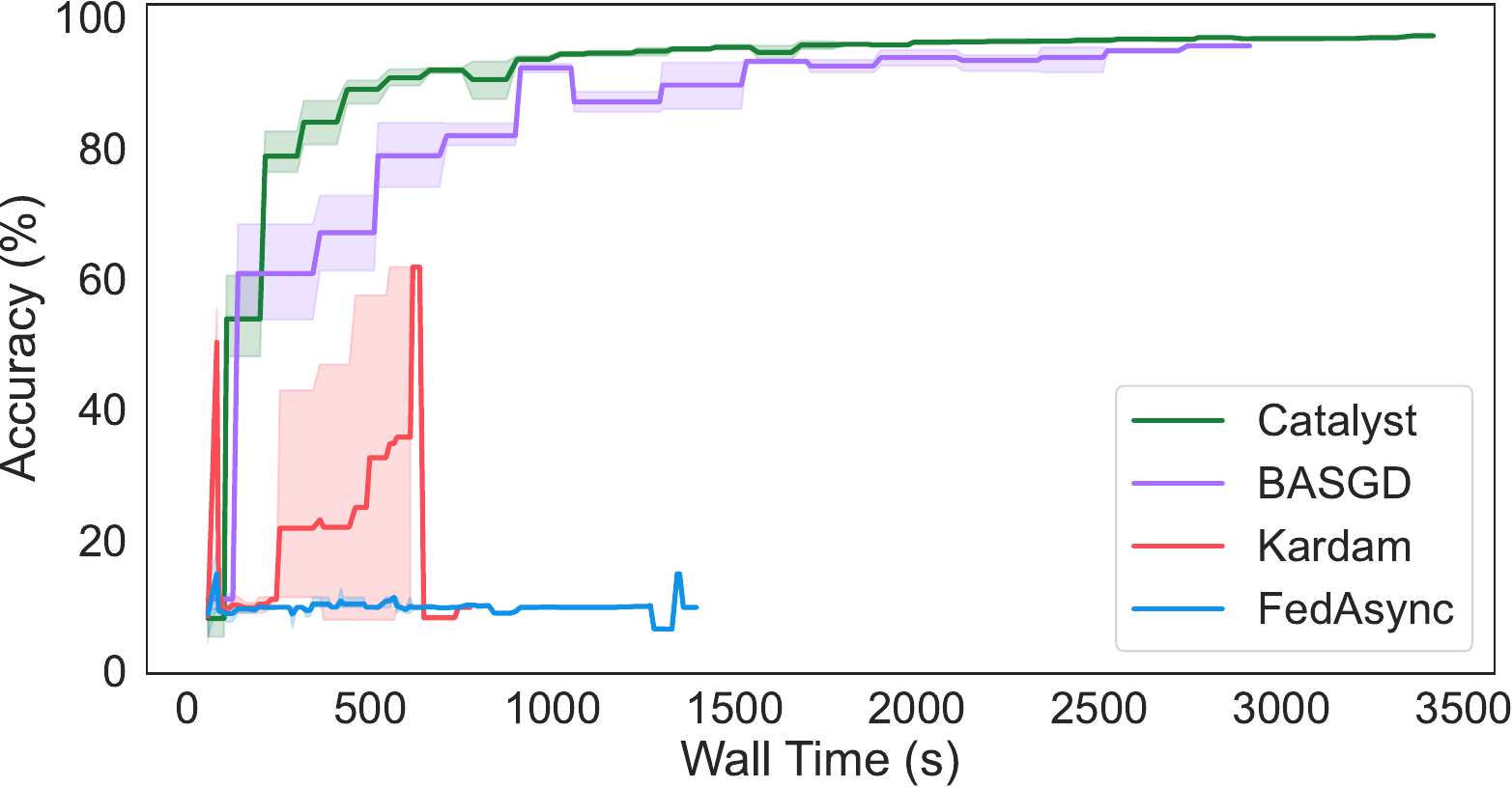}
		\caption{Accuracy with \mnist{} with the Gradient Inversion attack}
		\label{fig:acc_mnist_gi_atk}
	\end{subfigure}
	%comment
	%comment
	%comment
	%comment
	%comment
	%comment
	%comment
	\caption{Accuracy of the \pname{}, \texttt{Kardam}, \basgd{}, and \fasync{} defenses with the \mnist{} dataset without attack and with the Gradient Inversion attack.}
	%comment
	\label{fig:evaluations_byz_mnist}
\end{figure*}

Figure~\ref{fig:evaluations_byz_mnist} shows the two scenarios for \mnist{}. Without attack (Figure~\ref{fig:acc_mnist_no_atk}), the trends from Table~\ref{tab:time-perf} are reconfirmed, with the one note that \basgd{}, unlike the others, appears to struggle to converge. During an attack (Figure~\ref{fig:acc_mnist_gi_atk}), \pname{} maintains its speed of convergence, and still tops out at over 95\% accuracy, with \basgd{} improving considerably to claim the runner-up place. \texttt{Kardam} is unable to mitigate the attack for too long, and it also suffers from a large variance between runs, as its filtering mechanism proves unreliable. \fasync{} manages to run longer than \texttt{Kardam}, but to no avail, as it never makes any progress in terms of model quality.

\begin{figure*}[htp]
	\centering
	\begin{subfigure}[t]{0.45\textwidth} %comment
		\centering
		\includegraphics[width=\columnwidth]{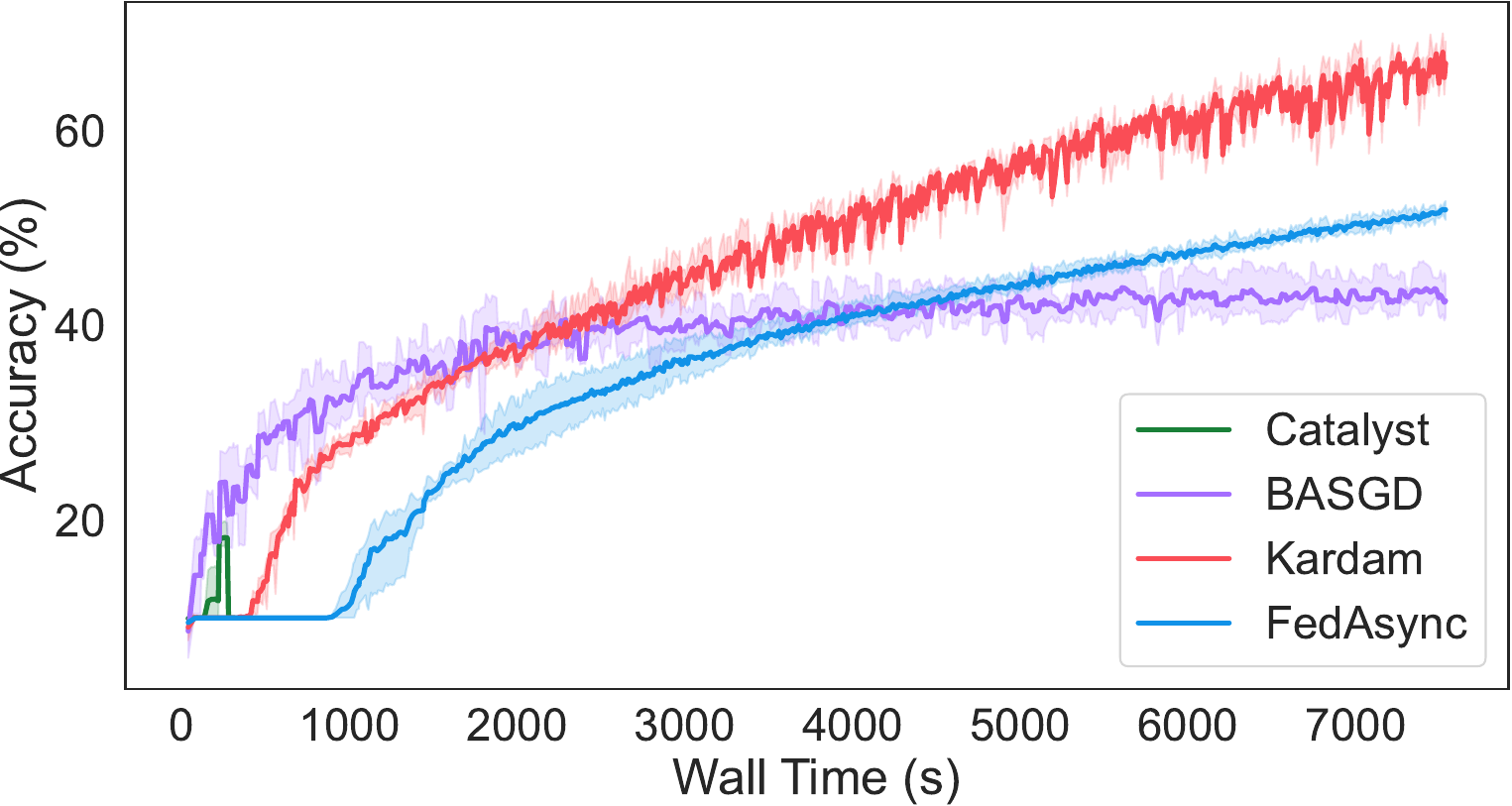}
		\caption{Accuracy with \cifar{} with no attacks}
		\label{fig:acc_cifar10_no_atk}
	\end{subfigure}
	\hfill
	\begin{subfigure}[t]{0.45\textwidth} %comment
		\centering
		\includegraphics[width=\columnwidth]{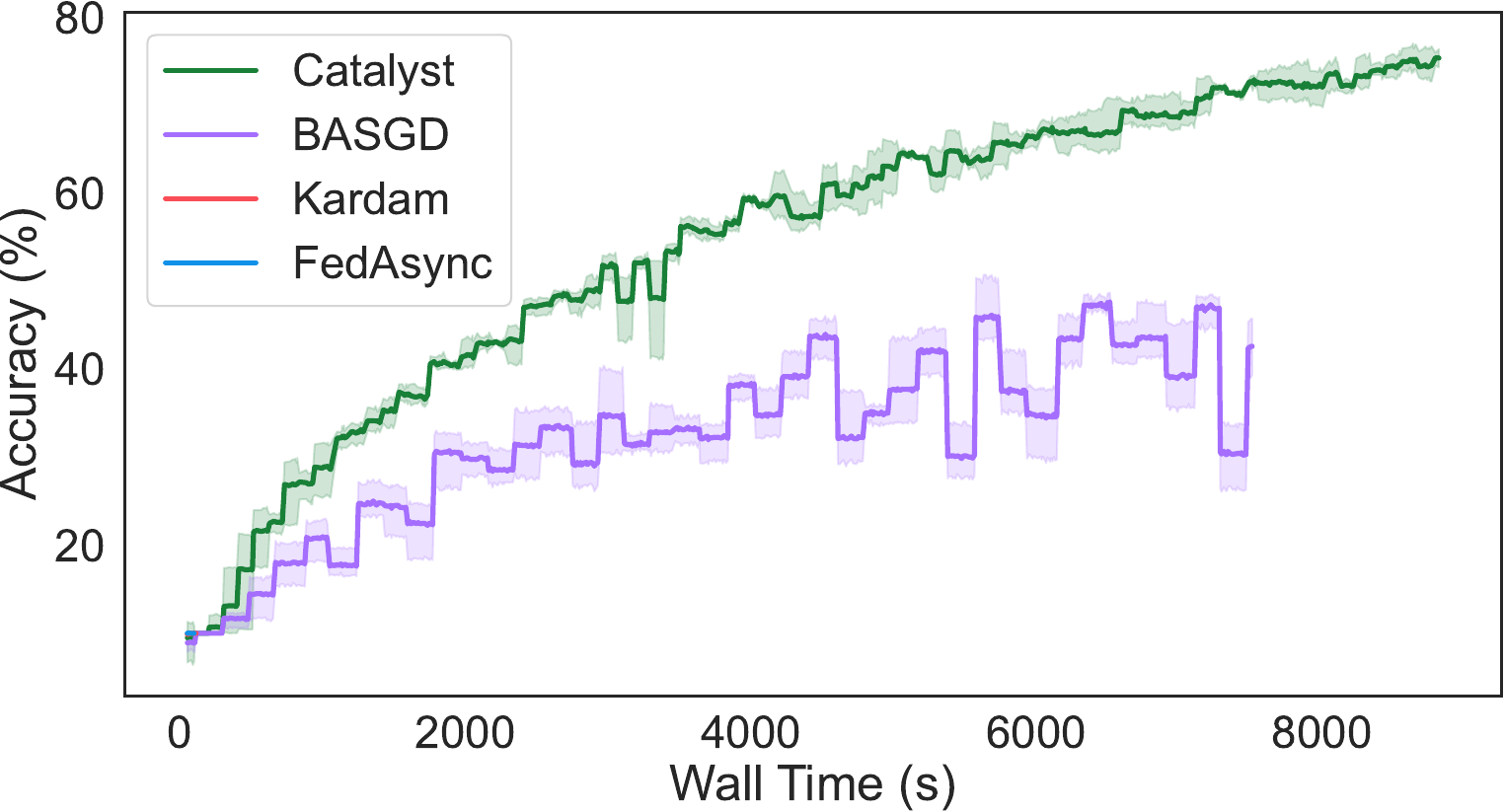}
		\caption{Accuracy with \cifar{} with the Gradient Inversion attack}
		\label{fig:acc_cifar10_gi_atk}
	\end{subfigure}
	%comment
	%comment
	%comment
	%comment
	%comment
	%comment
	%comment
	\caption{Accuracy of the \pname{}, \texttt{Kardam}, \basgd{}, and \fasync{} defenses with the \cifar{} dataset without attack and with the Gradient Inversion attack.}
	%comment
	\label{fig:evaluations_byz_cifar10}
\end{figure*}

Figure~\ref{fig:evaluations_byz_cifar10} concerns the \cifar{} image dataset and shows similar overall trends. Without attacks (Figure~\ref{fig:acc_cifar10_no_atk}), \texttt{Kardam} and \pname{} follow each other closely at the top, while \fasync{} continues to improve more slowly over time, and \basgd{} struggles to converge. Under attack (Figure~\ref{fig:acc_cifar10_gi_atk}), only \basgd{} and \pname{} achieve valid results for any period of time. Furthermore, they both manage to maintain their performance, although \basgd{} only tops out at around 40\% accuracy, while \pname{} reaches close to 80\%.

\begin{figure*}[htp]
	\centering
	\begin{subfigure}[t]{0.45\textwidth} %comment
		\centering
		\includegraphics[width=\columnwidth]{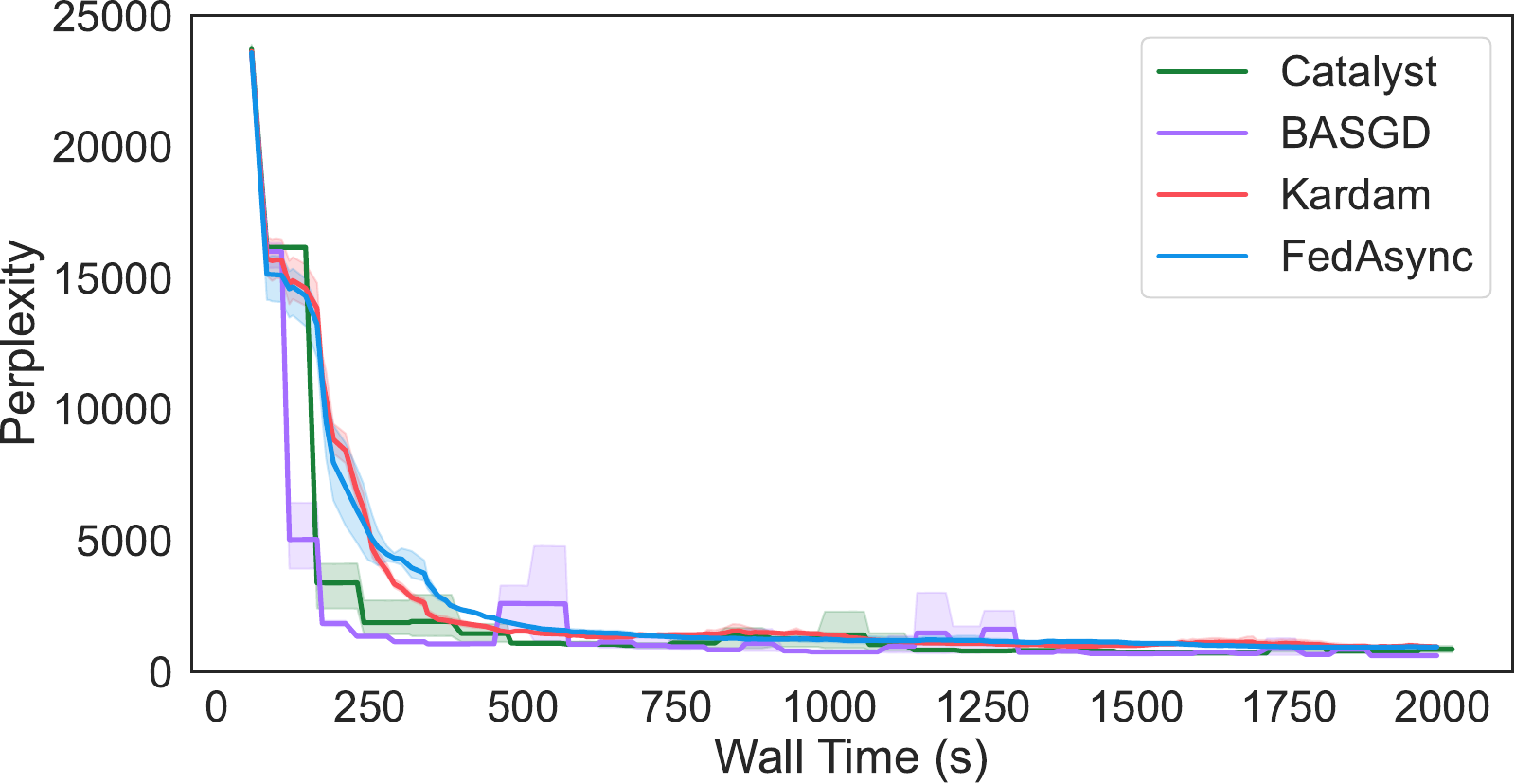}
		\caption{Perplexity with \wikitext{} with no attacks}
		\label{fig:acc_wiki2_no_atk}
	\end{subfigure}
	\hfill
	\begin{subfigure}[t]{0.45\textwidth} %comment
		\centering
		%comment
		\includegraphics[width=\columnwidth]{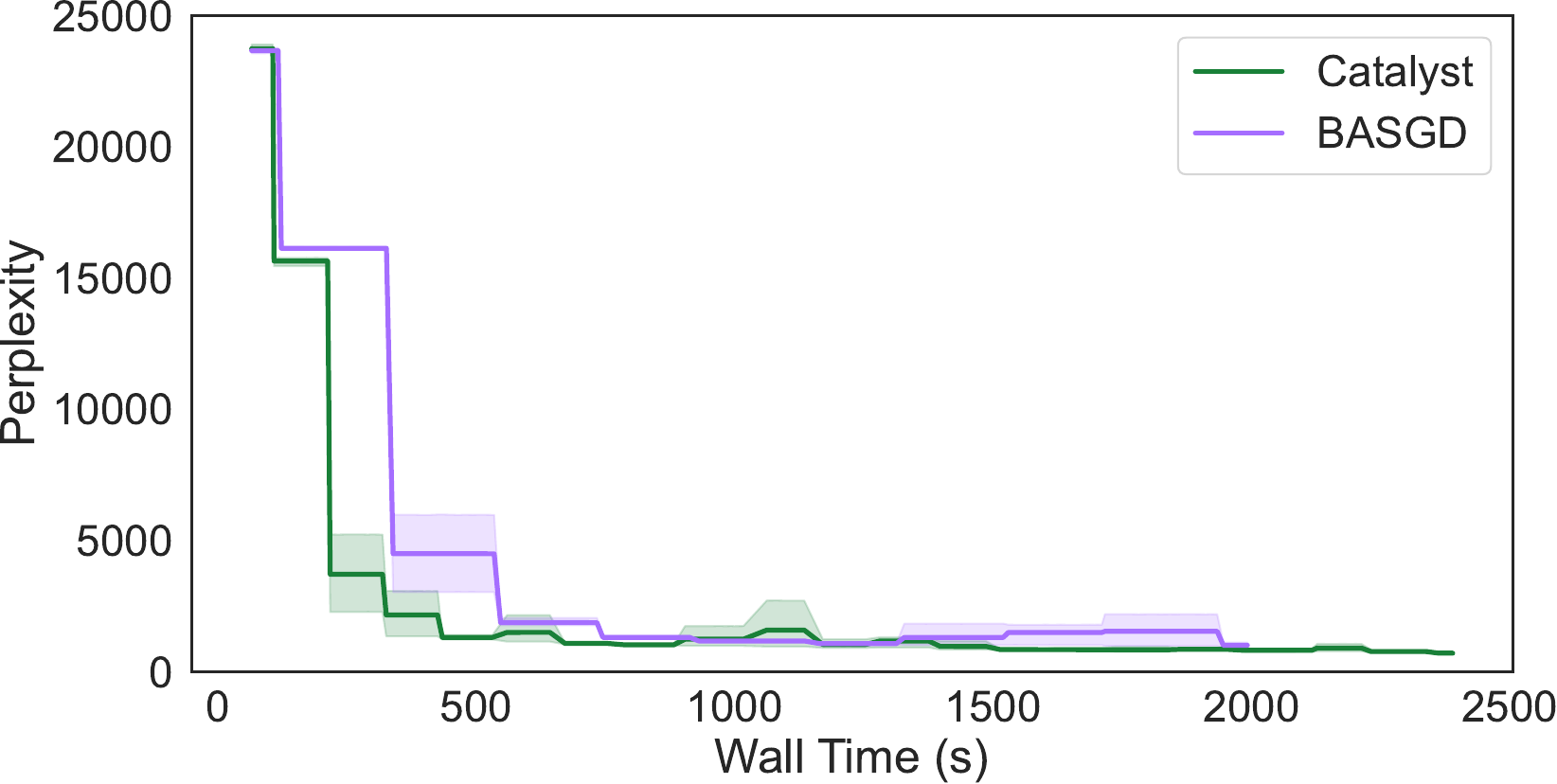}
		\caption{Perplexity with \wikitext{} with the Gradient Inversion attack}
		\label{fig:acc_wiki2_gi_atk}
	\end{subfigure}
	%comment
	%comment
	%comment
	%comment
	%comment
	%comment
	%comment
	\caption{Accuracy of the \pname{}, \texttt{Kardam}, \basgd{}, and \fasync{} defenses with the \wikitext{} dataset without attack and with the Gradient Inversion attack.}
	%comment
	\label{fig:evaluations_byz_wiki2}
\end{figure*}

For \wikitext{}, Figure~\ref{fig:evaluations_byz_wiki2} shows a closer match between the algorithms initially, but with little changes in the patterns observed later on. Thanks partly to the nature of the setting, where big strides are made in few steps, especially at the beginning, all methods perform well when no Byzantines are present (Figure~\ref{fig:acc_wiki2_no_atk}), even if \basgd{} and \pname{} end up being the leaders, as previously observed in the text setting. After enabling Byzantine clients (Figure~\ref{fig:acc_wiki2_gi_atk}), the effective methods remain the same two as before, and they also keep their relative ordering.

\subsection{Impact of the Number and Types of Clients}

Finally, we examine the robustness of our solution with different mixes of Byzantine clients $f$ in relation to the total client count $n$. We focus on the algorithms that attempt to protect against adversaries (\texttt{Kardam}, \texttt{BSGD} and \pname{}) and apply the harder gradient inversion attack again with the same strength as previously. We also choose server parameters based on the previously mentioned guidelines and repeat each experiment three times.

\begin{figure}[ht]
	\centering
	%comment
	%comment
	\includegraphics[width=\columnwidth]{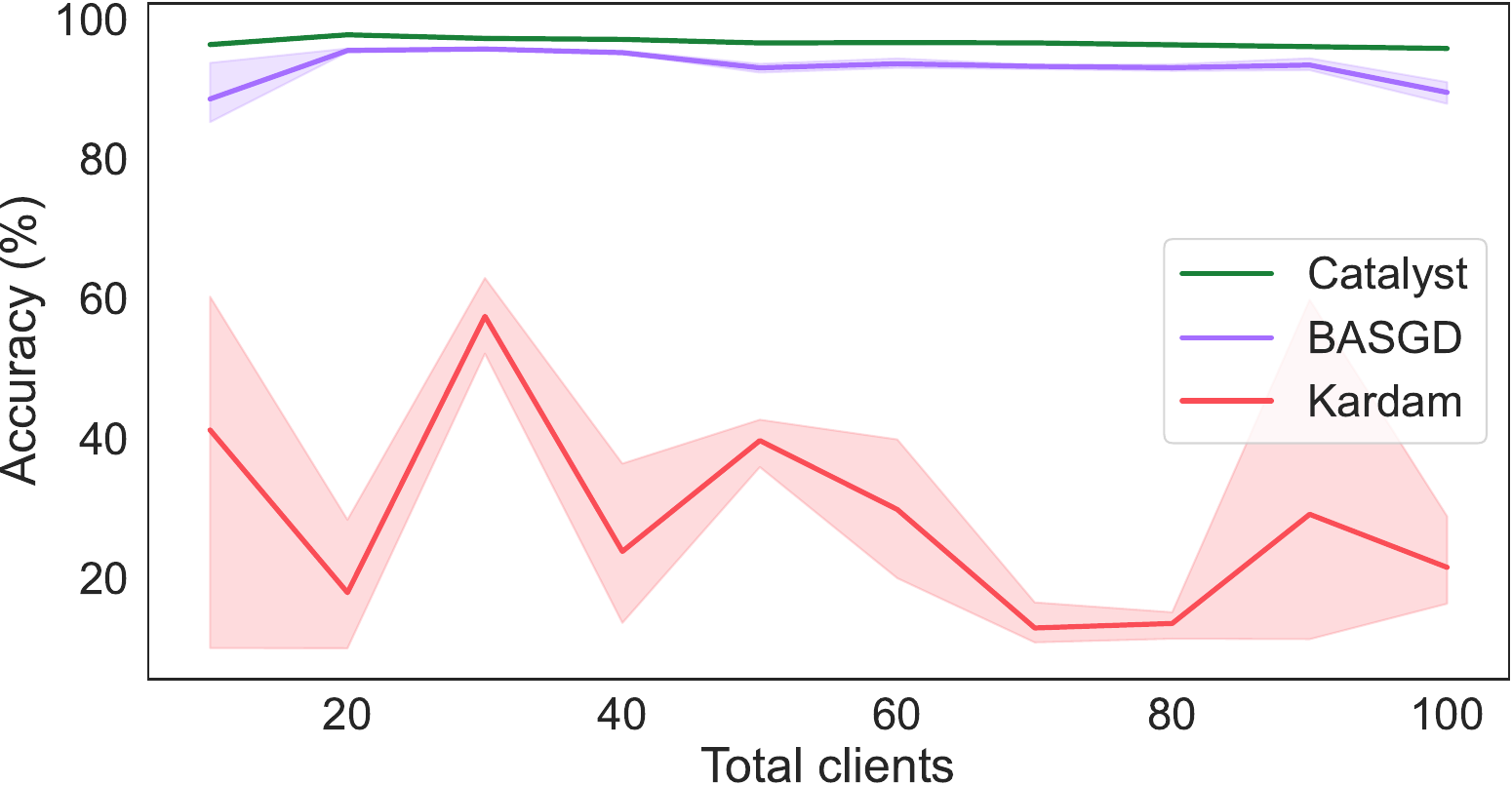}
	\caption{Accuracy depending on the number of clients with a fixed proportion of Byzantine clients (30\%) with \mnist{} under gradient inversion attack.
    }
	\label{fig:scalability_all}
\end{figure}

Figure~\ref{fig:scalability_all} showcases accuracy fluctuations for 10 to 100 total clients, while keeping the proportion of Byzantines at 30\%. As before, \texttt{Kardam} struggles to achieve reasonable performance regardless of $n$, and has very high variability between runs. \basgd{} also does not generally suffer from the fluctuations in $n$, and it achieves consistently good accuracy, of over 85\%. \pname{} improves upon \basgd{} by at least a couple of percentage points in all tests, especially in the extremes, with 10 and 100 clients, where its lead is even greater, especially as \basgd{} shows its largest variability for the small $n$.

\begin{figure}[ht]
	\centering
	%comment
	%comment
	\includegraphics[width=\columnwidth]{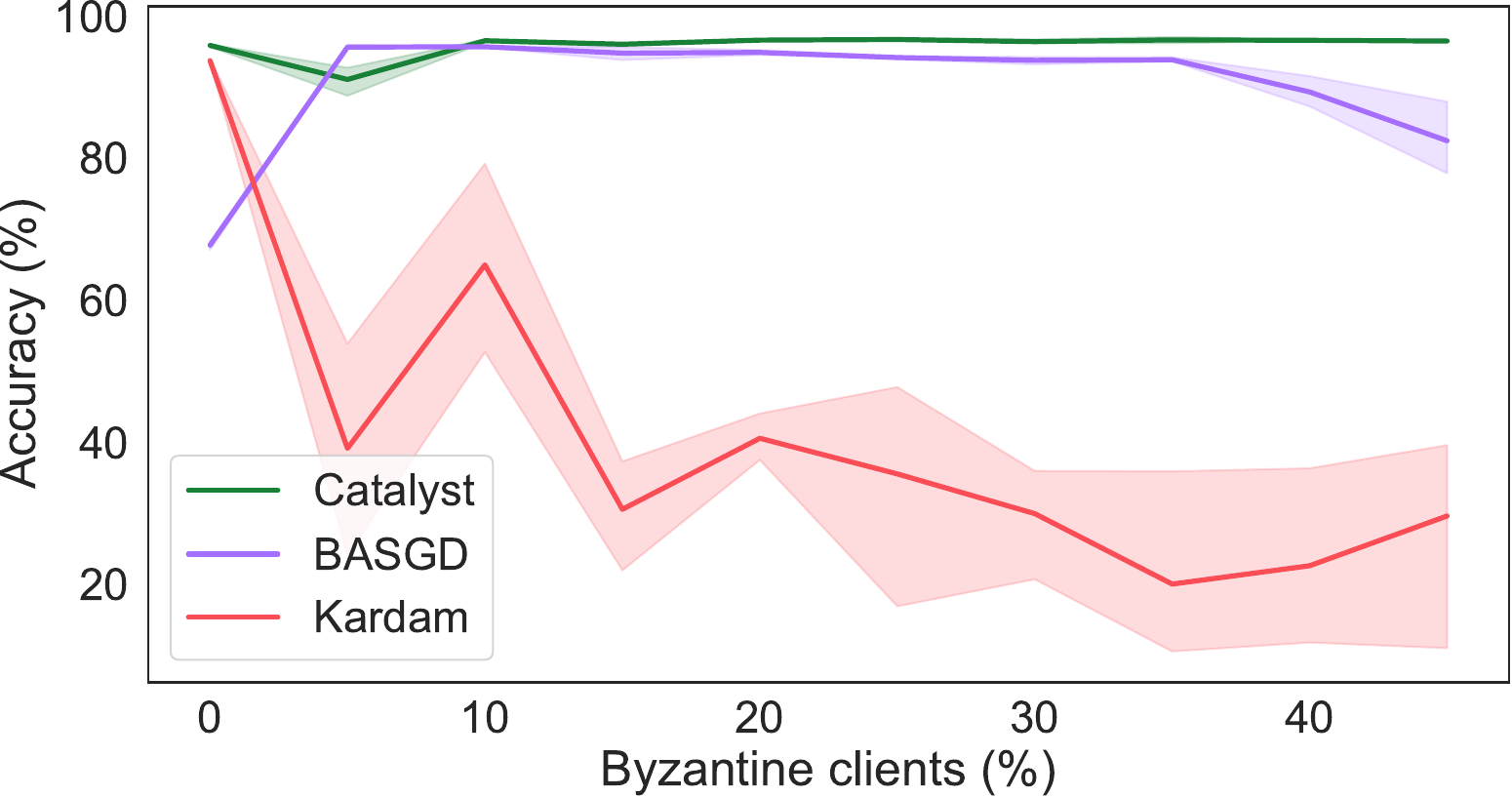}
	\caption{Impact of the proportion (\%) of Byzantine nodes with \mnist{} and 60 clients under gradient inversion attack.}
	\label{fig:scalability_byz}
\end{figure}

We highlight results when increasing the proportion of Byzantine clients in Figure~\ref{fig:scalability_byz} with a total $n=60$ clients. Here, \texttt{Kardam} starts strong, on par with its counterparts, but quickly degrades as we introduce more faulty nodes  and once again starts exhibiting large variability between repetitions. Our solution, \pname{} is insensitive to increases in the number of attackers. At the same time, \basgd{} once again falls behind it in the no-attack case, and when the ratio of attackers starts approaching 45\%.

\newpage

\section{Related work}
\label{sec:sota}

We analyse relevant previous works and compare \pname{} with relevant works in \autoref{tab:comparison_table}.

\begin{table*}[t]
	\centering
	\caption{Comparison of representative synchronous and asynchronous Byzantine Federated Learning algorithms. %comment
	}
	\label{tab:comparison_table}
	\rowcolors{2}{gray!10}{white}
	%comment
		\begin{tabular}{l|c|c|c|c}
			\rowcolor{gray!25}  & {\bf Async} & {\bf Does not require} &   &  \\
			\rowcolor{gray!25} \multirow{-2}{*}{\bf Method} &             & {\bf server dataset} & \multirow{-2}{*}{\bf Handles Stragglers} &  \multirow{-2}{*}{\bf Adversarial Model}                        \\ \hline

            FedAvg~\cite{mcmahan2017communication} & \xmark & \cmark & \xmark & \textcolor{BrickRed}{Fault-free} \\

            \fasync{}~\cite{xie2019fedasync} & \cmark & \cmark & \cmark & \textcolor{BrickRed}{Fault-free} \\

            FLAME~\cite{nguyenFLAMETamingBackdoors2022} & \xmark & \cmark & \xmark & \textcolor{OliveGreen}{$N > 2f$, Byzantine} but \textcolor{BrickRed}{no omission faults} \\
            
			Kardam~\cite{damaskinos2018asynchronous} & \cmark      & \cmark         & \cmark & \textcolor{OliveGreen}{$N > 3f$, Byzantine} \\
			
            Zeno++~\cite{xieZenoRobustFully2020}            & \cmark      & \xmark        & \cmark & \textcolor{OliveGreen}{$N > f$, Byzantine} \\
			BASGD~\cite{yangBASGDBufferedAsynchronous2021}  & \cmark      & \cmark & \xmark & \textcolor{OliveGreen}{$N > 3f$, Byzantine} \\
			\hline
           {\bf \pname} & \cmark & \cmark & \cmark  & \textcolor{OliveGreen}{$N > 3f$, Byzantine} \\ 
           \hline
           %comment
		\end{tabular}
	%comment
\end{table*}

\subsection{Synchronous FL}

McMahan et al. described the first Federated Learning (FL) frameworks, \texttt{FedSGD} and \texttt{FedAvg} in 2017~\cite{mcmahan2017communication}. 
In each round, the server selects a group of clients, send them its latest global model so that they can perform local training using their dataset. At the end of a round, the server collects the updates from all selected clients and aggregates them to update the global model. This training process is repeated until the global model converges. \texttt{FedSGD} and \texttt{FedAvg} eventually converge but may suffer from client heterogeneity as the server always wait for the slowest client. 

\subsection{Straggler Mitigation Techniques}

Several works attempted to mitigate the negative impact of stragglers in FL. 
The server can use a deadline for clients to submit their model updates, after which the server updates the global model and proceeds to the next round~\cite{li2019smartpc,nishio2019client}. In the same spirit, Bonawitz et al.~\cite{bonawitz2019towards} proposed for the server to select more clients than strictly necessary and only use the first received ones, which wastes client resources.  \texttt{FedProx}~\cite{li2018federated} allows clients to perform variable amount of work.  
\texttt{TiFL}~\cite{chai2020tifl} groups clients in tiers based on their performance characteristics, and selects a round's clients from a specific tier to reduce training times variance. 
Several works have designed systems where slow clients can partially offload their training tasks to faster clients~\cite{DBLP:conf/mass/DongZ0G20,ji2021computation,wu2021fedadapt,cox2022aergia}.

\subsection{Asynchronous FL}
In practical FL deployments, clients might have heterogeneous network bandwidths, computational power, and may not all be able to train a model simultaneously. Asynchronous FL systems address the straggler problem by allowing clients to participate in the global training at their own speed. In fully asynchronous FL systems, such as \fasync{}~\cite{xie2019fedasync}, the server modifies the global model with a client update as soon as it receives it and immediately returns the newly obtained global model to the client.   
Wu et al.~\cite{wu2021safa} describe several techniques, including improved model distribution, client selection and global aggregation, that can be used to improve the quality of the global model. 
\texttt{FedAT}~\cite{chai2020fedat} profiles client speed, classifies clients into tiers and combines synchronous intra-tier training and asynchronous cross-tier training. 
HFL~\cite{li2021stragglers} asynchronously pulls delayed local weights from stragglers. Additionally profile client speed. 
\texttt{ASO-Fed}~\cite{chen2020asynchronous} focuses on improving the accuracy of asynchronous FL in the case where clients have non-IID datasets.  Van Dijk et al.~\cite{vandijk2020asynchronous} decrease the amount of communication and add differential privacy, but they assume homogeneous clients. 
\texttt{FedBuff}~\cite{nguyen2022federated} updates the server's global model after receiving $K$ client updates, which enables the use of secure aggregation to preserve the privacy of client updates against an honest-but-curious server. The server buffer is implemented using cryptography or a trusted execution environment.
These asynchronous FL systems however do not tolerate Byzantine clients. In this work, we defend against Byzantine clients and maintain an asynchronous training process. 

\subsection{Byzantine Attacks on Federated Learning}

Malicious clients may launch untargeted attacks where they aim at degrading the accuracy of the global model, or targeted attacks where they aim for the global model to mislabel some inputs. 
In a label flipping attack~\cite{biggio2012poisoning,zhang2020adversarial}, a faulty client flips the labels of its dataset items to generate faulty gradients. 
In a Gaussian attack~\cite{fang2020local}, faulty clients send model updates that are sampled from a Gaussian distribution to the server. The \texttt{Krum}~\cite{blanchardMachineLearningAdversaries2017} and \texttt{Trim}~\cite{pmlr-v80-yin18a} attacks aim to push the global model in the opposite direction of correct gradients. Backdoor attacks~\cite{bagdasaryan2020backdoor,hongyi2020attack} aim at leading the global model to misclassify inputs in which a trigger has been inserted.   
In this work on focus on preventing model poisoning attacks. We however note that privacy attacks have been described in the literature and refer the reader to Jere et al.'s attack taxonomy for a more exhaustive discussion of these attacks~\cite{jere2021taxonomy}.

\subsection{Byzantine-Resilient Synchronous FL}

    %comment
    %comment
    %comment
    %comment
    %comment
    %comment

Byzantine-tolerant SGD algorithms have mostly focused on the synchronous and independent and identically distributed (IID) settings. These methods cannot be easily extended to asynchronous FL systems. 
\texttt{Krum}~\cite{blanchardMachineLearningAdversaries2017} updates the global model with the client update whose sum of Euclidean distances with other updates is the smallest. \texttt{GeoMed}~\cite{chenDistributedStatisticalMachine2017} uses a batch normalized median. \texttt{Trim-mean} and \texttt{Median}~\cite{pmlr-v80-yin18a} instead use a coordinate-wise median.
\texttt{Medoid}~\cite{xie2018generalized} also follows \texttt{Krum}'s idea but selects the client update to use based on the geometric median, margin median or median-around-median.  \texttt{GeoMed}~\cite{chenDistributedStatisticalMachine2017} and \texttt{Bulyan}~\cite{mhamdiHiddenVulnerabilityDistributed2018} generate an estimated global model update from the client updates.

Alistarh et al.~\cite{alistarhByzantineStochasticGradient2018} use historical information to identify harmful gradients. \texttt{DRACO}~\cite{chenDRACOByzantineresilientDistributed2018} utilizes coding theory and majority voting to recover correct gradients. 
\texttt{FABA}~\cite{xia2019faba} uses the Euclidean distance to remove outliers. The same authors later used multi-dimensional mean and standard deviation~\cite{xia2021defenses}. 
\texttt{Zeno}~\cite{xieZenoDistributedStochastic2019}, \texttt{Siren}~\cite{guo2021siren} and \texttt{ToFi}~\cite{xiaByzantineTolerantAlgorithms2023a} assume that the server holds a representative dataset that it uses to detect malicious client updates.  Cao and Lai~\cite{caoRobustDistributedGradient2018} propose robust synchronous SGD algorithms that rely on clients to raise alarms that the server uses to detect attacks. 
\texttt{FLTrust}~\cite{cao2020fltrust} uses a trust bootstrap to defend against Byzantine attacks. \texttt{FLAME}~\cite{nguyenFLAMETamingBackdoors2022} makes it possible for the server to detect malicious updates by clustering them. \texttt{SAFELearning}~\cite{zhangSAFELearningSecureAggregation2023} uses secure aggregation and detects a malicious update by statistically analyzing the magnitude of the model parameter vector. In this work, we take inspiration from clustering-based approaches and design an asynchronous Byzantine-resilient FL algorithm.   

\subsection{Byzantine-Resilient Asynchronous FL}

    %comment
    %comment
    %comment

Few FL algorithms attempt to tolerate Byzantine clients in asynchronous FL systems. \texttt{Kardam}~\cite{damaskinos2018asynchronous} identifies suspicious learning information during the training process using two filters. This approach works well even when there are communication delays between workers, however, when faced with a malicious attack, it has been shown that \texttt{Kardam} may also discard correct gradients, making it vulnerable to such attacks. 
\texttt{Zeno++}~\cite{xieZenoRobustFully2020} requires the server to use a representative auxiliary dataset, which might pose a privacy risk. 

In \basgd{}~\cite{yangBASGDBufferedAsynchronous2021}, the server maintains several buffers and assigns each client's model update to one of them. Notably, the buffer is reassigned only when a timer exceeds a predefined threshold, distinguishing it from the technique of bucketing, which undergoes reshuffling in every iteration. When all buffers accumulate enough non-empty updates, the server computes the average of model updates within each buffer, subsequently calculating the median or trimmed-mean of these averages to update the global model effectively. \basgd{} is sensitive to stragglers as the updates contained in a bucket are not aggregated until enough updates have been received. 
\texttt{AFLGuard}~\cite{fangAFLGuardByzantinerobustAsynchronous2022} presents a solution where the server checks if the client's model update moves in a similar direction as its own update and if the sizes of the updates are roughly the same. If both conditions are met, the server accepts the update and uses it to update the global model. Otherwise, it rejects the update and keeps the current global model unchanged. \texttt{AFLGuard} requires the server to have a data set available on the server that is representative of the data set that is used by the clients.
\texttt{EIFFeL}~\cite{roy2022eiffel} supports secure aggregation of verified updates, enforcing integrity checks and removing malformed updates without compromising privacy. Our algorithm, \pname{}, does not require the server to hold a trusted reference dataset, and we demonstrate that it trains a model faster than previous baselines and with higher accuracy. 

\section{Conclusion}
\label{sec:conclusion}

In this paper, we describe \pname{}, the first asynchronous and clustering-based Byzantine-resilient Federated Learning algorithm. \pname{} updates a global model every time $2f+1$ client updates that were computed on it are collected. These updates are clipped, clustered and finally filtered. The identified benign updates are then used to update the model and generate the next global model. As \pname{} is asynchronous, late updates from slow clients that relate to an old global model can be received. In that case, if these updates are not updated, then they are compared to previously received updates that are computed on the same global model and if they are found to be benign then they are integrated with staleness in the next global model computation. Our experiments on two image datasets (i.e., \mnist{} and \cifar{}), and one language dataset (i.e., \wikitext{}) demonstrate that \pname{} is faster than previous synchronous resilient algorithms and trains a model with higher accuracy than previous asynchronous resilient algorithms.  

\bibliographystyle{ACM-Reference-Format}
\bibliography{biblio}

\end{document}